%% file: main.tex
\documentclass[10pt,twocolumn]{article}
\usepackage[margin=0.75in]{geometry}
\usepackage[hyphens]{url}
\usepackage{graphicx}
\usepackage{natbib}
\usepackage{caption}
\usepackage{booktabs}
\usepackage{multirow}
\usepackage[acronym]{glossaries}

\glsdisablehyper
\newacronym{cot}{CoT}{Chain-of-Thought}
\newacronym{sql}{SQL}{Structured Query Language}
\newacronym{nl}{NL}{Natural Language}
\newacronym{scs}{SCS}{Structural Complexity Score}
\newacronym{bird}{BIRD}{BIg Bench for Large-Scale Database Grounded Text-to-SQL}
\newacronym{gee}{GEE}{Generalized Estimating Equations}

\title{Visible Reasoning Is Not a Universal Optimizer:\
Persona- and Thinking-Dependent Effects in Analytics Code Generation}
\author{
\begin{tabular}{c@{\qquad}c@{\qquad}c}
Bhawani Shankar Leelar & Pawan Chorasiya & Davin Hill \\
\textit{Optum AI} & \textit{Optum AI} & \textit{Optum AI} \\
\texttt{\scriptsize bhawani.leelar@optum.com} &
\texttt{\scriptsize pawan\_chorasiya@optum.com} &
\texttt{\scriptsize davin\_hill@optum.com} \\
\end{tabular}\\[0.8em]
\begin{tabular}{c@{\qquad}c}
Robert E. Tillman & Tamer Soliman \\
\textit{Optum AI} & \textit{Optum AI} \\
\texttt{\scriptsize rob.tillman@optum.com} &
\texttt{\scriptsize tamer.soliman@optum.com}
\end{tabular}}
\date{}

\begin{document}
\maketitle

\begin{abstract}
Visible \gls{cot} is often treated as a broadly useful reasoning
instruction, yet analytics code generation combines natural-language ambiguity,
schema grounding, target-language constraints, and model-specific inference
behavior. Because the same analytics request can be expressed in two distinct
target languages-SQL and Python (pandas)-this setting provides a natural
test of a common but under-examined assumption: that visible reasoning is more
effective when its representation matches the requested target, as in ``think
in SQL'' or ``think in Python.'' Together with generic instructions such as
``think step-by-step,'' such recommendations remain insufficiently evaluated
under controlled, execution-based comparisons. We study a query-matched SQL-pandas benchmark that crosses persona phrasing,
target language, visible-\gls{cot} format, control prefixes, direct generation,
and internal-reasoning configurations. The results do not support either a
universal accuracy advantage from visible \gls{cot} or a consistent benefit
from matching the reasoning representation to the target language. Instead,
the effects depend on the persona, target, model configuration, and
internal-reasoning setting. The control ablations further distinguish effects
of reasoning content from those of prompt format. These findings indicate that
reasoning strategies should be selected jointly for the model, persona, target,
and internal-reasoning configuration rather than adopted as universal
defaults. More broadly, the study provides a controlled framework for
identifying when visible reasoning improves executable generation, when it
primarily perturbs model behavior, and when the internal-reasoning
configuration is the more consequential factor.
\end{abstract}

\section{Introduction}
\label{sec:introduction}

Modern language models increasingly translate natural-language data requests
into executable analytics code. The same request can be realized in two different paradigms:
\gls{sql}, which specifies relational operations declaratively, and Python
(pandas), which expresses the result through a sequence of DataFrame
transformations. Visible \gls{cot} is now a common prompting device for both,
whether as a general instruction to reason step-by-step
\citep{wei2022chain} or as a target-matched instruction such as
\emph{think in SQL} or \emph{think in Python}. If the most useful visible-CoT representation is
the one that matches the requested code, it should preferentially benefit
procedural pandas generation. A query-matched SQL-pandas benchmark makes this proposition testable:
the database task, schema, question intent, and executable reference result can
be fixed while the requested target and visible-CoT representation vary.

Several alternative mechanisms complicate this target-matching hypothesis. A visible \gls{cot} may alter generated
code without supplying a better task solution: code-like notation can prime
target vocabulary, a fixed-step output contract can alter decoding behavior,
and any additional prefix changes the context presented to the model. A fluent
rationale is not necessarily a faithful account of the computation leading to
the final answer \citep{turpin2023language,lanham2023measuring}; additional
inference-time computation is also not uniformly useful across tasks
\citep{snell2025scaling}. We therefore distinguish two channels. In this paper,
\emph{visible CoT} (or \emph{visible reasoning}) denotes text in the externally
observable prompt-response exchange: a condition-specific CoT instruction or
prefix and the decoded \texttt{thinking} list returned with the final code. The
explicit-CoT conditions request eight labeled entries; direct generation and
the controls have different contracts described in
Section~\ref{sec:design}. This visible text is not taken to reveal latent
computation. \emph{Internal reasoning}
denotes unobserved model- or service-level inference-time computation. The
provider-exposed switches studied here are therefore termed
\emph{internal-reasoning configurations}; we neither observe nor interpret
provider-side internal traces \citep{zelikman2024quiet,hao2024coconut}. Thus,
toggling an internal-reasoning configuration changes an inference setting while
we continue to measure only visible output traces and executable code.

We examine these channels on a query-matched SQL-pandas \citep{li2023bird,bird_persona_2026} panel that crosses
persona phrasing, target language, visible CoT format, direct generation,
three control prefixes, and internal-reasoning configurations. The controls are
ablations of coherent NL \gls{cot}, not candidate prompting methods: length filler
removes task-relevant semantics while retaining approximate prefix volume;
scrambling disrupts word order while retaining related lexical material; and
wrong content supplies a reasoning-like prefix from a different task. Each is
compared with both direct generation and coherent NL \gls{cot}. This dual comparison
separates a change caused by loss of task-aligned reasoning from a change caused
by adding a long, structured context.

The central result is not a universal visible-\gls{cot} or target-matching advantage.
Instead, outcomes vary by persona, target, model configuration, and
internal-reasoning setting. In particular, the ablations show that reasoning
content and prompt format need not have the same behavioral role: a scrambled
prefix can match or exceed coherent NL \gls{cot} in localized cells even though it
does not preserve a chronological solution plan. Visible CoT also adds
token and latency overhead whose return is configuration-specific. The relevant
unit of comparison is consequently a model-persona-target configuration, not
a global prompting default.
In summary, this work makes four contributions:
\begin{itemize}
  \item We present a query-matched evaluation that holds the database task and
  executable reference result fixed while varying SQL or pandas targets,
  persona phrasing, direct generation, visible-CoT format, and model
  configuration.
  \item We test visible NL-CoT with three complementary controls: length filler,
  scrambled tokens, and wrong content. Each control is compared with both
  coherent NL-CoT and direct generation to separate task alignment from
  structured-prefix effects.
  \item We find that neither visible CoT nor target-language matching provides a
  universal accuracy advantage. The observed changes instead depend on the
  persona, target, and model configuration, with a localized scrambled-prefix
  result motivating a prompt-format explanation rather than a new default.
  \item We examine visible CoT and internal-reasoning configurations alongside difficulty,
  token use, and latency. These analyses show why direct generation remains an
  essential reference when selecting a reasoning configuration.
\end{itemize}

The next Section situates this design within prior work on visible CoT,
code-generation intermediates, and inference-time computation before specifying
the matched experiment.

\section{Related Work}

\paragraph{Visible \gls{cot} and causal interpretation.}
\gls{cot} prompting established that decoded intermediate text can improve
performance on multi-step tasks \citep{wei2022chain}. In prior work, such text
may be supplied in demonstrations, elicited as an output, or used alongside
unreported model-side computation; it should not automatically be read as an
internal reasoning trace. What an externally visible rationale contributes
remains less settled. Ablations of demonstrations and noisy rationales indicate
that relevance, ordering, and task fit can matter separately from local
plausibility \citep{wang2023understanding,zhou2024can}.
Other analyses document settings in which an incomplete or misleading trace
harms performance \citep{zheng2025the,su2026broken}. Faithfulness work further
cautions that a fluent explanation may not be the causal computation responsible
for the final answer \citep{turpin2023language,lanham2023measuring}. Our
length-filler, scrambled, and wrong-content conditions instantiate these
concerns as dual-baseline ablations of a visible NL-\gls{cot} prefix and output
contract for executable analytics generation.

\paragraph{Reasoning representations for code.}
Several code-generation approaches use plans, programs, or structured
intermediate representations to decompose a task.  Program-aided language
models delegate suitable subcomputations to an executable program
\citep{gao2023pal}, while work on intermediate languages and code prompting
compares natural-language, pseudocode, and code-like representations
\citep{deng2024assessing,sun2024unicoder,puerto2024code}.  Recent studies extend
this line to multilingual structured traces \citep{jin2025mscot} and to broad
comparisons of CoT strategies in programming benchmarks
\citep{jin2025understandingcotcode}. These contributions seek more effective
generation procedures and may use reasoning representations differently from
our prompt-response trace. In contrast, our objective is diagnostic: the task,
reasoning-slot structure, and evaluator are held fixed while the externally
visible representation, target language, and user phrasing are crossed.

\paragraph{Text-to-SQL planning and analytics generation.}
Reasoning is also central to text-to-SQL systems, commonly through schema
linking, query planning, verification, or repair.  SQL-of-Thought, for
example, combines staged planning and correction in a multi-agent procedure
\citep{chaturvedi2025sqlofthought}; STaR-SQL uses reasoning-oriented learning
and verification for text-to-SQL \citep{he2025starsql}.  Such systems change several ingredients jointly, including the number of calls, feedback, and selection policy. Our factorial design does not estimate the value of full agentic systems with extended planning, tool use, or open-ended multi-turn revision. Instead, it isolates whether the surface form of an otherwise matched visible prefix changes SQL or pandas execution accuracy under a bounded recovery policy.

\paragraph{Internal computation and adaptive inference.}
Quiet and latent reasoning approaches distinguish model-side computation from
the public token stream \citep{zelikman2024quiet,hao2024coconut}, and
test-time-compute work shows that additional inference effort should be
considered jointly with task difficulty and resource cost
\citep{snell2025scaling}. In code generation, uncertainty-guided \gls{cot}
similarly allocates extended reasoning selectively rather than
unconditionally \citep{zhu2025uncertcot}. These approaches
motivate our separate treatment of visible CoT and internal-reasoning
configurations, as well as the difficulty-conditioned token and latency
analyses. They do not license an equivalence between a longer visible trace and
provider-side internal reasoning; our configuration comparisons observe
behavioral consequences, not the content of internal computation.

\paragraph{Benchmarks, schema grounding, and persona.}
Text-to-SQL benchmarks such as Spider and BIRD foreground cross-domain schema
grounding and execution correctness \citep{yu2018spider,li2023bird}; data-science
benchmarks demonstrate the additional challenges of realistic Python-library
code \citep{lai2023ds1000,ouyang2025dsbench}. The matched setting used here
combines database-grounded requests and executable SQL and pandas targets
derived from BIRD \citep{li2023bird} with Original, Novice, and Expert persona
phrasings supplied by BIRD-Persona \citep{bird_persona_2026}. The paired
text-to-Python/text-to-SQL
setting makes persona a particularly important moderator, because the same
underlying intent can leave different amounts of procedural and schema detail
implicit \citep{hu2026benchmarking,ding2026ambisql}.  We therefore analyze
wording as part of the experimental design rather than averaging it away.

Taken together, this literature motivates a matched evaluation that separates
reasoning representation, task phrasing, and inference configuration. The next
Section defines that design and the execution-based estimands used to evaluate
it.

\section{Experimental Design}
\label{sec:design}

This Section defines the matched factors, common output contract, and
execution-based estimands used throughout the paper. It first specifies what is
held constant across conditions, then describes the model configurations and
the paired statistical comparisons that support the Results in
Section~\ref{sec:results}.

\subsection{Matched Panel and Conditions}

From the 1,534-task BIRD-Persona benchmark \citep{bird_persona_2026}, seeded
random sampling retained at most 10 questions from each database-difficulty
stratum. After the pre-inference exclusions documented in Appendix~\ref{sec:appendix-reproducibility}, 304
database-grounded questions form the common panel for all six completed model
configurations. For every
question, database schema and executable gold answer are fixed while phrasing
(Original, Novice, Expert), target (SQL, pandas), and reasoning condition vary.
The four explicit \gls{cot} conditions
share the same eight semantic slots. The representation contract, not the
decomposition, is manipulated within those four conditions.
\begin{description}
  \item[NL.] Complete each slot with concise English prose and avoid formal
  code bodies.
  \item[SQL.] Use clause, projection, source, join, predicate, aggregate,
  subquery, alias, and validation fragments.
  \item[Python.] Use literals, assignments, selectors, masks, pandas
  transformations, and assertions.
  \item[Pseudocode.] Use an uppercase structured vocabulary including
  \textsc{Output}, \textsc{Map}, \textsc{Tables}, \textsc{Columns},
  \textsc{Compute}, \textsc{Filter}, \textsc{Build}, \textsc{Convert}, and
  \textsc{Validate}.
  \item[Direct.] Receive no visible-CoT definition and return an empty visible
  thinking list.
\end{description}
Three ablations of coherent NL-CoT alter properties of its visible prefix:
\begin{description}
  \item[Length filler.] Ablate task-relevant NL CoT sentences while retaining
  approximately matched textual volume in an eight-label unrelated prefix.
  \item[Scrambled.] Ablate chronological order by permuting the words of a
  reasoning-like eight-step template while retaining its lexical material.
  \item[Wrong content.] Ablate task relevance by supplying a reasoning trace
  associated with a different query matched by database/difficulty where
  possible.
\end{description}

Thus each model configuration contains
$304\times3\times2\times8=14{,}592$ matched trials. Appendix~\ref{sec:appendix-protocol} specifies
the prompt slots, controls, query exclusions, evaluator, and full condition
matrix.

Every trial uses a shared schema-grounded JSON response contract containing a
visible \texttt{thinking} list, target-language code, and an explicit code-type field. The four
explicit-CoT prompts require exactly eight labeled steps with the same semantic
roles and differ only in the required representation; direct generation requires
an empty visible thinking list. The controls retain the shared code and schema
guardrails but replace the coherent NL-CoT prefix with the relevant ablation.
Thus, the exact-eight-step statistic reported in the Results is an observed
response-compliance measure, not a substitute for execution accuracy or a
measurement of latent reasoning. Internal-reasoning configurations are toggled separately; they change the inference setting, while the study continues to measure only visible output traces and executable code. Full prompt and control construction details are given in Appendix~\ref{sec:appendix-protocol}.

The task panel supplies a database-grounded request, a schema
context, executable SQL and pandas reference programs, and three semantically
matched persona phrasings for the same analytic intent
\citep{bird_persona_2026}. Original is the benchmark wording from \citep{li2023bird}; Novice
uses a less technical user-facing formulation; Expert makes analytic details
more explicit.  We retain the benchmark's paired formulation rather than
treating these as different questions.  Consequently, a within-query
comparison isolates a change in prompt wording or reasoning condition while
holding the database, reference output, and target language fixed.  This is
especially important for SQL-pandas comparison: the same analytic result can
be declarative in SQL yet require an explicit sequence of DataFrame operations
in pandas.

Appendix~\ref{sec:appendix-annotations} defines the benchmark-derived annotations used in the adjusted
analyses: BIRD difficulty, an AmbiSQL-derived balanced ambiguity score
\citep{ding2026ambisql}, a SQL-grounded phrasing-specificity score, and
\gls{scs}, an equal-weight SQL-structure count.

\begin{table}[t]
\centering
\begin{tabular}{lp{0.58\columnwidth}}
\toprule
Model & Full evaluated model and configuration \\
\midrule
GLM4 & GLM-4-32B-0414; non-reasoning \\
Qwen3 & Qwen3-30B-A3B-Instruct-2507; non-reasoning \\
Gemma4-off & Gemma 4 26B-A4B-IT; internal reasoning disabled \\
Gemma4-on & Gemma 4 26B-A4B-IT; internal reasoning enabled \\
Gemini Flash-0 & Gemini 2.5 Flash; internal-reasoning budget 0 \\
Gemini Flash-24k & Gemini 2.5 Flash; internal-reasoning budget 24,576 \\
\bottomrule
\end{tabular}
\caption{Six completed configurations used in the experiments.}
\label{tab:model-settings}
\end{table}

The 304 questions are present for every completed model configuration and every
factorial cell.  The common frame prevents a model-specific missing row from
changing the query distribution used in a paired contrast; Appendix~\ref{sec:appendix-reproducibility}
reports the pre-comparison audit that produced this frame.
Table~\ref{tab:model-settings} gives the full names and establishes the compact
reference names used in the remainder of the paper.

The open-weight configurations were served with vLLM on an AWS
\texttt{g6.12xlarge} instance using BF16 weights and a 32,768-token context
limit. Gemini Flash was queried through the provider API from an Apple M4 Max
client with 64~GB unified memory; provider-side inference hardware was not
observable. Full infrastructure details, including completion limits, retries,
timeouts, software dependencies, and the execution environment, are reported in
Appendix~\ref{sec:appendix-serving}. These serving paths motivate within-configuration,
rather than cross-provider, resource comparisons.

\subsection{Evaluation and Inference}

Generated code is executed in an isolated worker and matched to a freshly
executed reference result after deterministic normalization. Parse failures,
runtime failures, and missing outputs are incorrect. The primary estimand is
the paired within-query accuracy change relative to direct generation. Exact
two-sided McNemar tests, 4,000 query-cluster bootstrap intervals, and
Benjamini-Hochberg (BH) correction evaluate 252 primary
model-persona-target-condition contrasts. Declared secondary families cover
internal-reasoning configuration contrasts, target matching, persona
interactions, and the control mechanism tests. A complementary query-clustered
logistic \gls{gee} adjusts for
target, BIRD difficulty, database, AmbiSQL-derived ambiguity \citep{ding2026ambisql}, SQL-grounded
phrasing specificity, and \gls{scs} \citep{hu2026benchmarking}; exact paired contrasts remain primary. Complete procedures and all
correction families are reported in Appendix~\ref{sec:appendix-statistics}. Each completed cell uses
one zero-temperature generation under the fixed experiment seed; the analysis
therefore estimates conditional behavior of these six serving configurations,
not sampling variability over repeated stochastic decodes.

For a reasoning condition $m$ and direct generation $0$, the paired effect is
the query-average difference
\[
\widehat{\Delta}_m = \frac{1}{n}\sum_{q=1}^{n}
\left(y_{qm}-y_{q0}\right),
\]
reported in percentage points. Here $n=304$, $q$ indexes the retained query, $m$ is the
reasoning condition being compared with direct generation, $0$ denotes direct
generation, and $y_{qm}\in\{0,1\}$ is the exact-execution correctness indicator
for query $q$ under condition $m$. Exact McNemar tests use only discordant query
pairs and are appropriate because both conditions are evaluated on the same
underlying task.  The 95\% intervals resample complete query clusters rather
than individual rows, preserving the dependence induced by the two targets,
three phrasings, and eight conditions attached to each query.  BH correction
controls the declared family of 252 model-persona-target-condition tests.
The same-family internal-reasoning, target-language matching, persona
interaction, and
control-mechanism questions have separate, explicitly reported correction
families because they answer different estimands.

The complementary \gls{gee} is deliberately not a replacement for paired inference.
It asks whether a condition association persists after including model configuration,
target, persona, difficulty, database, ambiguity, phrasing specificity, and
\gls{scs}, with query-clustered robust covariance.  This model supplies a
check on confounding by annotated query properties; it does not turn correlated
benchmark features into independently randomized causes.  Token and latency
summaries are similarly descriptive.  They retain individual \gls{cot}
formats, condition on difficulty, and compare values within a serving
configuration rather than asserting that latency is directly comparable across
infrastructure.

This design preserves query-level pairing while making the scope of each
comparison explicit. The next Section reports the execution
accuracy, ablation, internal-reasoning, and resource findings.

\section{Results}
\label{sec:results}

This Section first establishes direct generation as the execution reference,
then evaluates visible-CoT formats, ablations, internal-reasoning configurations, and
resource distributions against that reference. The paired-inference design from
Section~\ref{sec:design} is retained in every comparison.

\subsection{Direct Generation as the Reference}

Table~\ref{tab:original-detailed-condition-accuracy} gives the corresponding
descriptive results for Original phrasing. Direct generation is an essential
part of this comparison rather than a residual baseline: its accuracy differs
substantially by target and model configuration. Across the six configurations,
direct SQL accuracy exceeds direct Python accuracy by 8.22-16.45 percentage
points, and direct generation ties the strongest direct-or-explicit-CoT value
for Gemma4-off on Python. The four explicit formats do not produce a common
ranking across SQL and Python. Together, these cells
establish a target- and configuration-dependent reference against which the
paired tests below should be read.

\subsection{No Universal Visible-\gls{cot} Advantage}

No condition survives the 252-test primary family, and no target-matched
format has a BH-corrected advantage (Table~\ref{tab:inference-summary}).
The direct-reference matrix above is therefore descriptive evidence of
heterogeneity, not a replacement for the corrected paired tests.

\begin{table}[t]
\centering
\begin{tabular}{lrr}
\toprule
Inference family & Tests & BH discoveries \\
\midrule
Mode-direct & 252 & 0 \\
Target matching & 36 & 0 \\
Internal-reasoning toggle & 96 & 16 \\
Persona interactions & 756 & 1 \\
\bottomrule
\end{tabular}
\caption{Multiplicity-controlled inference families. Primary condition tests
are the first row; other rows address distinct predeclared questions.}
\label{tab:inference-summary}
\end{table}

Persona nevertheless matters. The paired persona-interaction family in
Table~\ref{tab:inference-summary} and the all-persona \gls{gee} both retain
mode-by-persona evidence after BH correction, indicating that phrasing is not a
uniform shift in task difficulty. \gls{scs} does not retain a distinct adjusted
association in that model, whereas ambiguity and phrasing specificity do; these
covariates describe the benchmark rather than causal persona mechanisms. The
appendix supplies the complete persona-by-difficulty matrices, which show why
the Original-focused main table must not be generalized to every phrasing
regime. It also shows that a control can be the highest cell in a row,
particularly under Novice wording. This pattern motivates the dual-baseline
ablation analysis below instead of a post-hoc selection of the highest-scoring
prefix.

\begin{table*}[t]
\centering
\begin{tabular}{llrrrrrrrrr}
\toprule
\multirow{2}[3]{*}{Model} & \multirow{2}[3]{*}{Target} & & \multicolumn{5}{c}{CoT} & \multicolumn{3}{c}{Control} \\
\cmidrule(lr){4-8} \cmidrule(lr){9-11}
 &  & Direct & NL & SQL & Python & Pseudocode & mean & Filler & Scrambled & Wrong \\
\midrule
GLM4 & SQL & 23.68 & 28.29 & 28.29 & 25.99 & \textbf{29.28} & 27.96 & 24.01 & 25.66 & \textbf{27.30} \\
GLM4 & Python & 15.13 & \textbf{19.74} & 18.42 & 17.11 & 17.76 & 18.26 & \textbf{16.45} & 15.13 & 15.79 \\
Qwen3 & SQL & 29.93 & \textbf{33.55} & 30.26 & 33.55 & 30.59 & 31.99 & 31.58 & \textbf{34.54} & 30.26 \\
Qwen3 & Python & 21.71 & \textbf{22.04} & 21.38 & 20.39 & 20.72 & 21.13 & 23.03 & \textbf{25.33} & 24.01 \\
Gemma4-off & SQL & 36.18 & 37.17 & \textbf{37.83} & 37.50 & 36.51 & 37.25 & \textbf{37.17} & 36.84 & 36.51 \\
Gemma4-off & Python & \textbf{19.74} & 17.76 & 19.41 & \textbf{19.74} & 18.09 & 18.75 & 18.42 & 18.42 & \textbf{18.75} \\
Gemma4-on & SQL & 37.83 & 37.83 & 38.16 & 38.49 & \textbf{39.14} & 38.40 & 39.14 & 39.47 & \textbf{41.12} \\
Gemma4-on & Python & 26.32 & 27.30 & \textbf{27.63} & 26.32 & 25.66 & 26.73 & 24.34 & \textbf{25.99} & \textbf{25.99} \\
Gemini Flash-0 & SQL & 35.86 & 36.51 & 37.17 & 35.53 & \textbf{39.80} & 37.25 & 39.47 & 39.14 & \textbf{41.12} \\
Gemini Flash-0 & Python & 23.03 & 25.00 & 25.99 & 22.04 & \textbf{26.32} & 24.84 & \textbf{26.97} & 25.00 & 24.34 \\
Gemini Flash-24k & SQL & 36.84 & \textbf{37.50} & 35.86 & \textbf{37.50} & 36.51 & 36.84 & 36.18 & \textbf{36.51} & 33.22 \\
Gemini Flash-24k & Python & 26.32 & 25.00 & \textbf{27.63} & 25.99 & 26.32 & 26.23 & 26.64 & \textbf{27.63} & 23.36 \\
\bottomrule
\end{tabular}
\caption{Original-phrasing exact execution accuracy (\%) along with the mean of the four explicit-CoT formats. In each row, bold
marks the best direct-or-explicit-CoT value and the best control value.}
\label{tab:original-detailed-condition-accuracy}
\end{table*}

The paired accuracy comparisons must be read with their resource distribution
in view. Figures~\ref{fig:original-toggle}, \ref{fig:original-token}, and
the appendix latency analysis retain target, difficulty, and individual CoT
formats: the internal-reasoning view uses signed enabled-minus-disabled resource
buckets, whereas the token view locates direct and visible-CoT accuracy in
model-adaptive bins. These views show that visible CoT usually occupies larger
token and latency budgets, but the associated accuracy return is small and
configuration-specific rather than systematic. Harder queries shift both direct
and prompted generations toward larger budgets, and enabling internal reasoning
changes the resource range occupied by paired evaluations. Thus, the results do
not support visible CoT as a generally cost-effective default; they support
configuration-specific selection against a direct-generation reference.

\subsection{Controls as NL-\gls{cot} Ablations}

The controls test semantic relevance, order, and task alignment in a visible
prefix; they are diagnostic ablations, not optimized prompts.
Table~\ref{tab:original-control} compares each control with both coherent
\gls{nl} \gls{cot} and direct generation. Length filler and wrong content show
no corrected difference from either comparator. Scrambling shows no corrected
difference from coherent \gls{nl} \gls{cot}, but is modestly higher than direct
generation. This does not imply that scrambled reasoning is useful. Rather, it
suggests that a related but non-chronological prefix can sometimes alter the
model's generation regime without preserving a valid stepwise plan.

\begin{table}[t]
\centering
\setlength{\tabcolsep}{3pt}
\begin{tabular}{lrrl}
\toprule
Ablation & $\Delta$ NL & $\Delta$ direct & BH outcome (NL; direct) \\
\midrule
Length filler & -0.36 & 0.90 & not detected; not detected \\
Scrambled & 0.16 & \textbf{1.43} & not detected; higher \\
Wrong content & -0.49 & 0.77 & not detected; not detected \\
\bottomrule
\end{tabular}
\caption{Original-phrasing control contrasts, pooled over six configurations,
targets, and matched queries. $\Delta$ NL is control minus NL-CoT accuracy;
$\Delta$ direct is control minus direct-generation accuracy. Changes are
percentage points. ``Not detected'' denotes failure to reject after BH
correction; ``higher'' is an observed corrected
contrast.}
\label{tab:original-control}
\end{table}

\begin{figure}[t]
\centering
\includegraphics[width=\linewidth,height=0.35\textheight,keepaspectratio]{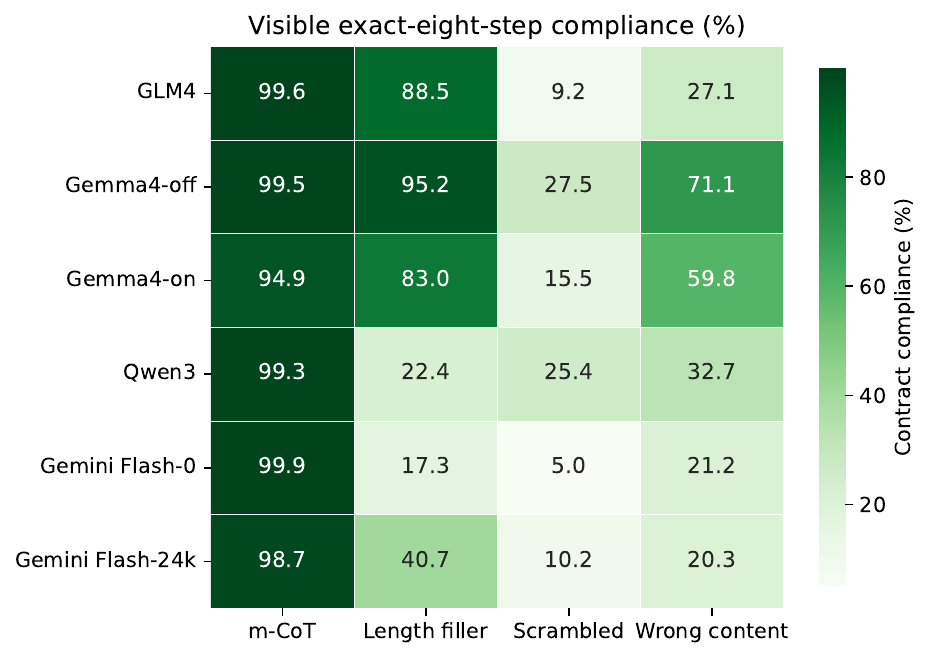}
\caption{Exact eight-step compliance by model. \emph{m}-CoT averages the four
explicit-CoT formats.}
\label{fig:control-compliance}
\end{figure}

The full matrix shows that this effect is persona- and difficulty-dependent,
with the strongest localized result under Novice-SQL. Figure~\ref{fig:control-compliance}
and the appendix compliance tests show why the effect should not be
attributed to recovery of the eight-step contract: scrambled outputs have low
exact-step adherence, and compliance is not associated with correctness within
the scrambled control. The supported interpretation is therefore a
configuration-specific sensitivity to perturbed task-related context, not a
general benefit of incoherent reasoning.

\subsection{Internal-Reasoning Configuration and Resource-Bounded Accuracy}

\begin{figure*}[t]
\centering
\includegraphics[width=\textwidth,height=0.72\textheight,keepaspectratio]{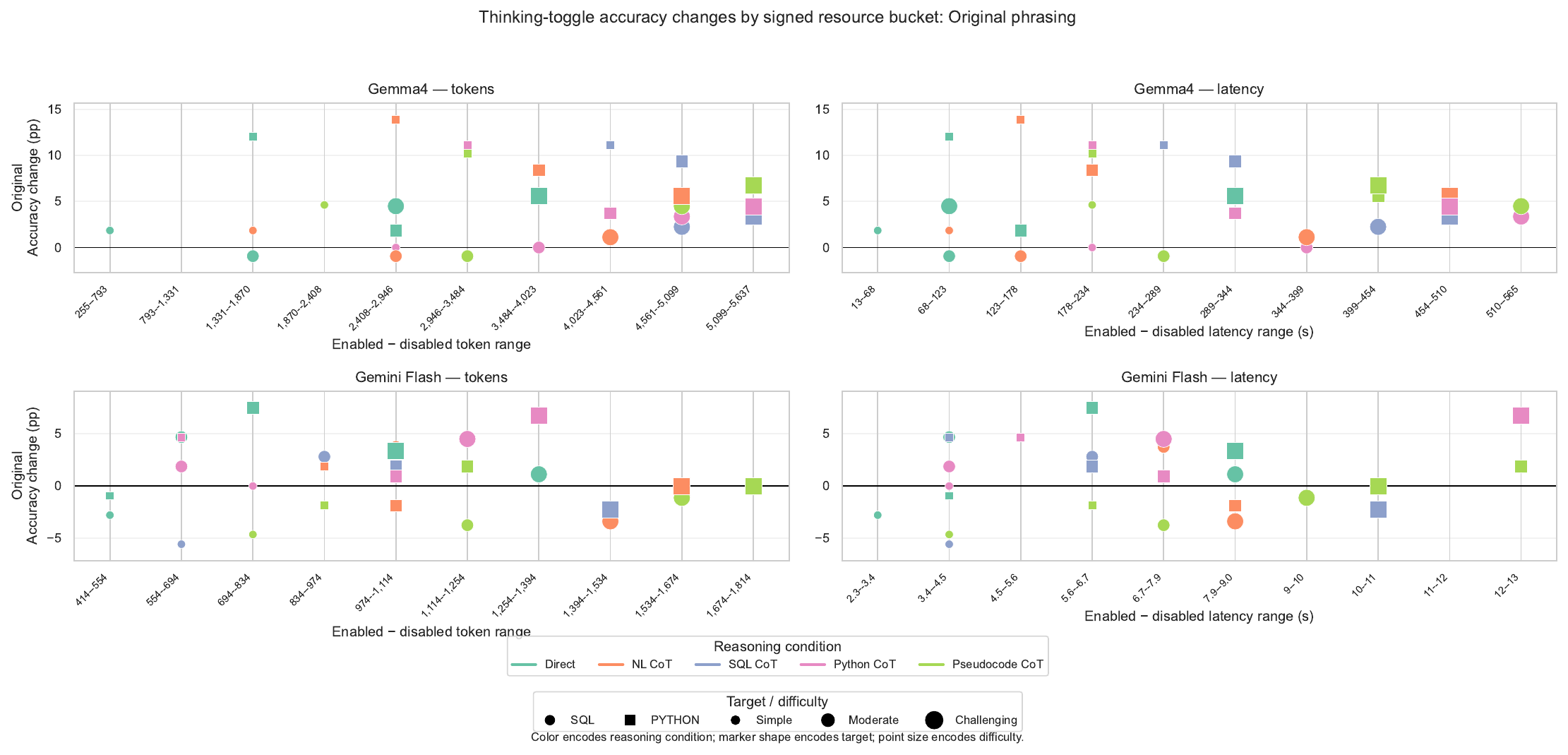}
\caption{Original-phrasing internal-reasoning configuration effects within
family-adaptive signed token and latency buckets. Each point is the
mean paired enabled-minus-disabled accuracy change in its resource bucket.}
\label{fig:original-toggle}
\end{figure*}

The signed resource plots in Figure~\ref{fig:original-toggle} provide the
model-family detail behind the preceding overview. The paired points occupy
different resource ranges for Gemma4 and Gemini Flash, and their accuracy
changes differ across targets and difficulty levels. Thus, enabling internal
reasoning redistributes evaluations across resource budgets as well as changing
accuracy; it cannot be interpreted as a common treatment across model families.
The complete
three-persona display is provided in the appendix.

\begin{figure*}[!h]
\centering
\includegraphics[width=\textwidth,height=0.72\textheight,keepaspectratio]{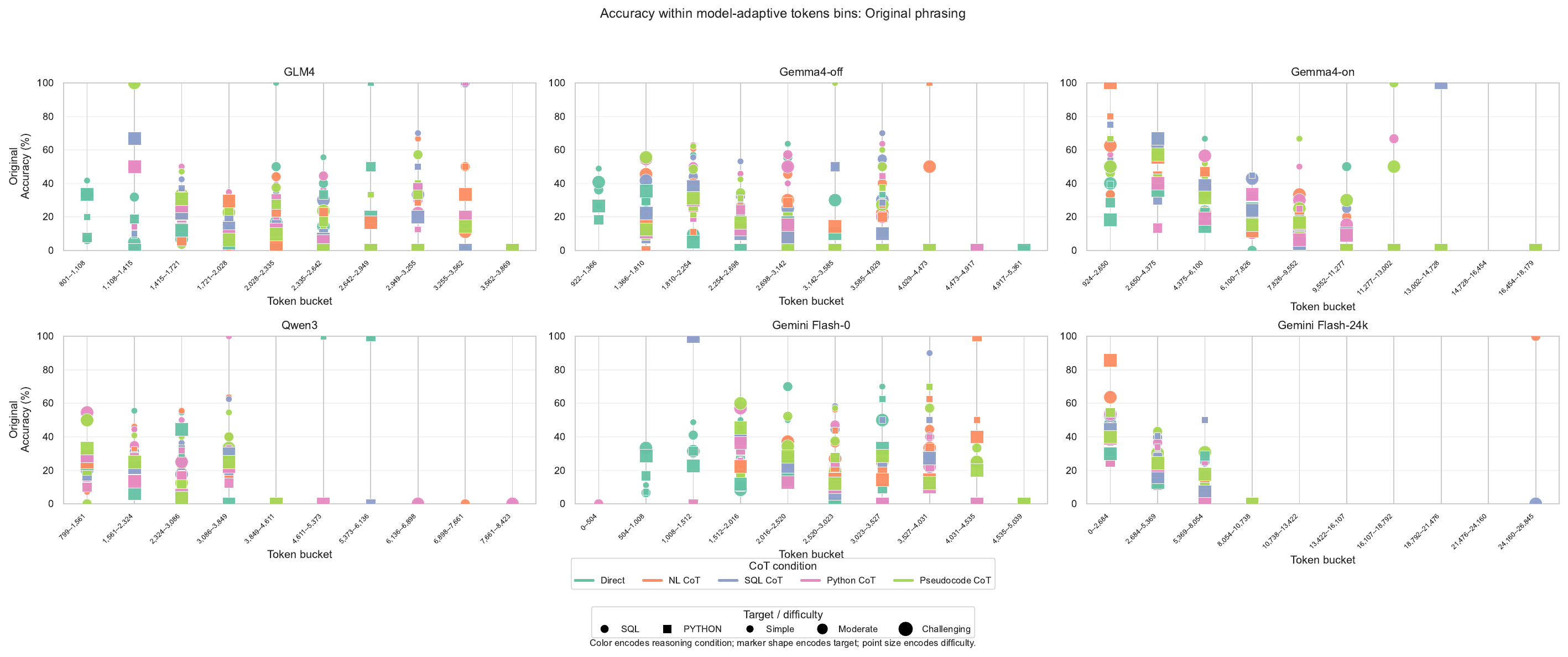}
\caption{Original-phrasing accuracy within model-adaptive total-token bins. Each point is the mean accuracy in a bin.}
\label{fig:original-token}
\end{figure*}

Visible \gls{cot} must also be evaluated against its resource cost.
Figure~\ref{fig:original-token} keeps the four formats separate and conditions on
target and difficulty. Direct and explicit-CoT points are distributed across
model-adaptive token bins rather than along a shared accuracy curve: harder
queries shift both target languages toward larger budgets, while visible CoT
generally extends that spread. The relative positions of direct and CoT points
differ by model and target, and latency view (in the appendix) should be
compared only within a serving configuration. These budget-stratified
summaries are descriptive rather than causal dose-response plots or
cross-provider latency comparisons.

These results distinguish the observed patterns from their causal
interpretation. The next Section considers the mechanism hypotheses and the
resulting evaluation guidance.

\section{Discussion}
\label{sec:discussion}

The results favor a conditional account of visible CoT. Direct generation
is not uniformly dominated: Table~\ref{tab:original-detailed-condition-accuracy}
shows that it can match the strongest direct-or-explicit-CoT option, while other
cells favor different explicit formats. The primary analysis therefore does not
support either a universal accuracy benefit or a target-language-matched
advantage for the tested panel. Persona is not an ornamental factor: the
adjusted interaction and the complete appendix matrices show that phrasing
changes the operating regime in which a prompt is evaluated.

The ablations sharpen the interpretation. Their results do not support the
claim that a longer prefix alone explains accuracy movements, nor do they show
that models reliably follow an injected wrong plan. Instead, the surprising
scrambled outcome is consistent with an interaction between task-related tokens
and a visible output structure. This is a falsifiable mechanism hypothesis: a
frozen-trace experiment can independently manipulate token content, sequence,
and step format to determine which component carries the effect.

For deployment, the relevant comparison is an accuracy-resource frontier, not a
single accuracy-per-token ratio. Figures~\ref{fig:original-toggle},
\ref{fig:original-token}, and the appendix latency analysis show that
visible \gls{cot} usually occupies larger token and latency budgets, while its
accuracy gain is small and configuration-specific. Difficulty also shifts both
direct and prompted generations toward larger budgets, and internal-reasoning
settings change the resource range of paired evaluations. These bucket views support
configuration-specific selection against direct generation, not a global
recommendation to add visible \gls{cot} or enable internal reasoning.

The next Section consolidates these conditional findings and their implications
for evaluating reasoning prompts in analytics-code generation.

\section{Conclusion}

In matched analytics-code generation trials, direct generation remains a
competitive, configuration-dependent reference rather than a uniformly weaker
alternative to visible \gls{cot}. Visible \gls{cot} is neither a universal optimizer
nor a reliably target-language-matched one; its effects vary with persona,
target, model configuration, and internal-reasoning configuration. The
dual-baseline ablations indicate that prompt-format and lexical effects can
perturb generation independently of coherent visible CoT, while the
difficulty-conditioned resource analyses show why configuration-specific
choices are necessary. Together, these findings provide a controlled basis for
future work on when visible CoT helps code generation, when it changes behavior
without improving it, and when an internal-reasoning configuration is the more
consequential lever.

\section{Limitations}

The controls are diagnostic ablations of coherent NL-\gls{cot}, not
counterfactual edits of one frozen rationale. They preserve different
properties of an injected prefix and therefore cannot separately identify the
causal roles of semantic content, sequence, length, and output structure. The
wrong-content and scrambled prefixes provide reasoning-like contexts, not
interventions on latent computation; the scrambled-control findings should be
read as mechanism hypotheses about sensitivity to perturbed task-related
context, not as evidence that incoherent reasoning is generally beneficial.

Each matched cell contributes one temperature-zero evaluation under a fixed
bounded-recovery policy. Because the runner permits limited recovery after
transient call failures, JSON-parse failures, and execution failures, the
outcome reflects a controlled single-task evaluation loop rather than either a
pure first-attempt sample or an unrestricted agentic system. Repeated decoding
seeds, service replicas, and independently sampled wrong-content traces would
be needed to quantify stochastic and operational variance. Persona phrasing,
ambiguity, phrasing specificity, difficulty, and \gls{scs} are correlated
benchmark descriptors, so the adjusted \gls{gee} associations should not be
read as independent causal effects.

The six configurations provide structured coverage, not a randomized model
sample. Same-family internal-reasoning contrasts are informative, but
architecture, training data, weights, access path, serving stack, and provider
implementation remain confounded across families. Token fields may be
provider-reported or locally estimated, and latency is client-observed
wall-clock time rather than a hardware-normalized service-time metric; resource
analyses therefore support within-configuration comparisons against direct
generation, not cross-provider cost benchmarks.

Finally, the task scope is executable SQL and pandas analytics generation under
a shared JSON response contract. The conclusions may not transfer to open-ended
software engineering, interactive tool-using agents, alternative execution
engines, multi-turn repair systems, or substantially different reasoning
interfaces.

\appendix
\input{appendix}

\input{main.bbl}
\end{document}

%% file: appendix.tex
\section{Factorial and Artifact Audit}

The common analysis frame contains 87,552 unique factorial cells: six model
configurations, 304 query IDs, three personas, two targets, and eight reasoning
conditions. Every configuration contributes 14,592 rows. There are no duplicated
setting-query-persona-target-condition keys, no missing final-correctness
labels, and no missing \gls{scs} values in the evaluated frame. The
internal-reasoning-enabled configurations use the same cached native gold
executions and common question set as their paired disabled or budget-zero
configurations.
Table~\ref{tab:supp-inventory} summarizes the completed common-frame inventory.

This appendix records the construction details needed to audit the annotations,
exclusions, controls, serving settings, and analyses.

\subsection{Annotation score definitions}
\label{sec:appendix-annotations}

The adjusted analyses use three benchmark-derived annotations: phrasing
specificity, ambiguity, and SQL structural complexity. They are not additional
experimental treatments. They are fixed query-level or phrasing-level
descriptors used to audit whether a reasoning-condition comparison is
confounded with properties inherited from the BIRD-Persona benchmark panel.

\paragraph{Phrasing specificity.}
Phrasing specificity is the benchmark-provided wording annotation used in the
adjusted analyses to reflect how explicitly a question wording states
schema-grounded details relative to its paired SQL task
\citep{bird_persona_2026}. We treat it only as a fixed inherited benchmark
descriptor of the displayed wording, not as an experimental treatment or a
stand-alone measure of semantic equivalence.

\paragraph{Ambiguity.}
Ambiguity is derived from an AmbiSQL-style ambiguity detector and taxonomy
\citep{ding2026ambisql}. Let $L_x$ be the multiset of level-2 ambiguity labels
detected for wording $x$, and let $K$ be the number of level-2 taxonomy labels
($K=7$ in the detector used here). The coverage and volume components are
\[
  A_{\mathrm{cov}}(x)=\frac{|\mathrm{unique}(L_x)|}{K},
  \qquad
  A_{\mathrm{vol}}(x)=\frac{|L_x|}{|L_x|+K}.
\]
The balanced ambiguity score used as the GEE covariate is
\[
  A_{\mathrm{bal}}(x)=
  \lambda A_{\mathrm{cov}}(x)+(1-\lambda)A_{\mathrm{vol}}(x),
\]
with $\lambda=0.5$. Thus the score lies in $[0,1]$, increases when more
ambiguity categories are detected, and also gives diminishing credit for
multiple detected ambiguity items. It is a study-specific metric derived from
AmbiSQL outputs, not a claim about AmbiSQL's original evaluation protocol.

\paragraph{Structural Complexity Score.}
\gls{scs} is an equal-weight count of SQL structural features computed from the
executable reference SQL after parsing it in the SQLite dialect. Following the
matched text-to-Python/text-to-SQL benchmark's use of \gls{scs}
\citep{hu2026benchmarking} and related text-to-SQL workload profiling practice
\citep{ma2025comparing},
let $c^{\mathrm{col}}_q$, $c^{\mathrm{tab}}_q$,
$c^{\mathrm{where}}_q$, $c^{\mathrm{join}}_q$, $c^{\mathrm{agg}}_q$,
$c^{\mathrm{scalar}}_q$, $c^{\mathrm{cte}}_q$, and
$c^{\mathrm{subq}}_q$ denote the counts of distinct referenced columns,
physical tables, atomic \texttt{WHERE} predicates, join pairs, aggregate calls,
nonaggregate scalar-function calls, common-table-expression definitions, and
nested subqueries for query $q$. Then
\[
  \begin{split}
  \mathrm{SCS}_q =
  c^{\mathrm{col}}_q+c^{\mathrm{tab}}_q+c^{\mathrm{where}}_q
  +c^{\mathrm{join}}_q\\
  +c^{\mathrm{agg}}_q+c^{\mathrm{scalar}}_q
  +c^{\mathrm{cte}}_q+c^{\mathrm{subq}}_q .
  \end{split}
\]
The score characterizes the reference SQL structure. A generated SQL or pandas
program can be correct without reproducing the same structure.

The following subsections distinguish the instructions imposed on a model from
response properties measured after generation.

\subsection{Prompt construction and compliance protocol}
\label{sec:appendix-protocol}

The main paper specifies the shared three-field JSON response and eight
semantic slots. At implementation level, SQL must be one self-contained SQLite
query. Pandas code executes in a supplied DataFrame environment, cannot import
modules or access files, networks, or processes, and must assign the answer to
\texttt{result}. Schema identifiers must come from the displayed context.
These target-specific guardrails are identical across reasoning conditions.

The three controls replace the coherent NL-CoT prefix while retaining the same
schema, target-code, and JSON contracts. Length filler is a seeded, unrelated
eight-step trace assembled from an everyday-task phrase bank. Its text is
extended until its count under the experiment's shared token counter
approximately matches a deterministic eight-step reasoning-like template for
the current query and target. Scrambling applies a seeded permutation to the
whitespace-delimited tokens of that template, preserving its token multiset but
removing sequence and its visible step labels. Wrong content selects a different
query with the same database identifier and difficulty when available, using a
seed determined by the query identifier. It injects that source query's
pre-generated NL-CoT trace from a fixed bank; if no bank entry is available, it
uses a deterministic non-gold eight-step trace derived only from the source
question, database identifier, and target. Neither path exposes reference SQL,
reference pandas code, or execution evidence to the target trial.
For the retained study bank, source-query NL-CoT traces were generated offline
with \texttt{gpt-5-mini}
(bank prompt version \texttt{v4}) and stored only as thinking-step text plus
source-query metadata; evidence fields, gold SQL, and gold pandas code were
excluded from the bank entries.

We distinguish instruction compliance from correctness. For explicit CoT, the
eight-string requirement is part of the prompt. For controls, the injected
prefix preserves, destroys, or replaces visible step labels but does not add
the explicit CoT conditions' exact-eight-output instruction. Exact eight-step
output is therefore an observed behavioral measure rather than a uniform
control requirement. We parse each response against the shared JSON contract
and record the number of generated visible thinking steps. Exact step compliance
is the indicator that the parsed thinking list contains eight strings. The main-paper
heatmap aggregates this indicator by model and reports explicit-CoT as its mean
over the four formats; the full format-level matrix is reported in
Figure~\ref{fig:supp-control-compliance}. Execution correctness remains the
primary outcome throughout.

\subsection{Reproducibility protocol}
\label{sec:appendix-reproducibility}

The seeded database-difficulty sampling described in the main paper produced
310 candidates. Before inference, five were excluded because native SQL gold
answers were not executable and one because its required function was
unsupported by the SQLite evaluator. Exclusions were fixed before condition
comparisons and did not depend on model output. Intersecting completed settings
then yielded the audited 304-query common frame in
Table~\ref{tab:supp-inventory}.

For each query, the database identifier, difficulty label, schema context,
question phrasing, target language, reasoning prefix, generated answer,
execution status, normalized output, correctness label, token counts, and
latency were retained.  The gold answer was executed once per database and
target under the same deterministic evaluator used for model outputs.  A model
output was marked correct only when its executed result matched the normalized
gold result; parse failures, runtime failures, and missing outputs were marked
incorrect rather than silently discarded.  No semantic judge was used for the
primary label.

Query order was fixed by the common manifest, and all summaries followed key
and missingness audits. Reported or locally estimated
token fields were retained: provider/serving-stack usage metadata was preferred
when present, with a common local counter used as fallback. Latency was measured
locally with a monotonic, high-resolution clock around the client
request/response path and accumulated with the timed trial phases; it is not
supplied by a provider. These
measurements support within-configuration runtime descriptions rather than a
hardware-normalized comparison across serving paths.

Generation used a fixed experiment seed (20260630) and temperature zero for
the completed runs. Each matched cell contributed one recorded generation; the
analysis does not average repeated decoding seeds. Internal-reasoning configurations
are identified in Table~\ref{tab:supp-inventory} and the main model table. They
are request or service configurations that may alter model-side inference, but
no provider-side thought tokens or internal-reasoning traces were requested,
retained, or analyzed; every configuration is evaluated through the same visible
response contract and executable code. Model-specific token budgets and
provider-side internal-reasoning controls are therefore treated as serving
configurations rather than as
architecture isolations. Authentication, transport retries, and provider response
metadata were handled before the final row-level audit; no failed request was
silently replaced by a second sampled question.

No reasoning format, correction family, or resource-bin width was selected by
maximizing accuracy. The model panel is descriptive coverage of completed
configurations, not a hyperparameter search.

\subsection{Serving and runtime configuration}
\label{sec:appendix-serving}

Table~\ref{tab:supp-serving} records the execution environment for the six
completed configurations. The open-weight configurations were served with vLLM on an
AWS \texttt{g6.12xlarge} instance provisioned with 48 vCPUs, 192~GiB memory,
and four NVIDIA L4 GPUs (96~GB aggregate GPU memory). Inference used BF16
weights and a 32,768-token maximum context length. The internal-reasoning switch
for Gemma4 was passed at request time, so Gemma4-off and Gemma4-on use the same
checkpoint while differing in the enabled inference channel. The Gemini Flash
configurations were queried
through the provider API from a Mac with an M4 Max processor and 64~GB unified
memory. This computer was an API client and experiment orchestrator; it was
not the provider's inference host. Accordingly, resource totals compare
conditions within a serving configuration and should not be read as a
hardware-normalized benchmark between the two serving paths.

The experiment configuration fixes temperature at zero, the experiment seed at
20260630, a request timeout of 600 seconds, at most two runner-level recovery
attempts after the initial generation (shared across transient-call, JSON-parse,
and execution repair), and a 30-second execution timeout.
Non-thinking vLLM generations use a 4,096-token completion limit. Gemma4-on
starts at 8,192 tokens and, only when the completion reaches that limit,
retries the same prompt at 12,288 and then 16,384 tokens subject to the
32,768-token context bound. Gemini Flash uses provider-side
internal-reasoning budgets of 0 and 24,576; no thought summaries are requested,
and the provider's final-output limit is not overridden. The analysis
analysis environment used Python 3.12 with pandas, SQLGlot, matplotlib,
seaborn, statsmodels, and vLLM 0.25.1. The original row-level records do not
preserve a complete operating-system image, CUDA/driver image, or provider-side
hardware manifest. Each latency is client-observed wall-clock time: it includes
the instrumented request/response call and other timed trial phases, rather
than an internal provider service-time field. It is therefore a
within-configuration outcome.

Open-model servers used a bounded context window to retain key-value-cache
headroom during batched serving. Generated programs were evaluated in an
isolated worker with deterministic normalization before comparison against a
fresh native reference execution. Per-request timeout, transport retry,
parsing, execution status, token fields, and locally measured end-to-end
latency were retained in the row-level artifact. A failed generation
contributes an incorrect outcome rather than causing a query to be resampled.

\begin{table}[t]
\centering
\begin{tabular}{lp{0.56\columnwidth}}
\toprule
Evaluation path & Runtime configuration \\
\midrule
GLM4; Qwen3 & vLLM on AWS \texttt{g6.12xlarge} (4 L4 GPUs, 48 vCPUs, 192~GiB); BF16; 32,768-token context; no internal-reasoning setting. \\
Gemma4-off/on & Same AWS/vLLM host and checkpoint; BF16; 32,768-token context; request-level internal reasoning disabled/enabled. \\
Gemini Flash-0/24k & Provider API, issued from Mac M4 Max with 64~GB unified memory; provider-side serving hardware not observed. \\
All configurations & Temperature 0; fixed experiment seed; bounded repair and execution; deterministic reference normalization. \\
\bottomrule
\end{tabular}
\caption{Serving and execution configuration. Full model identifiers and the
compact reference names used elsewhere are listed in the main model table.}
\label{tab:supp-serving}
\end{table}

\begin{table}[t]
\centering
\begin{tabular}{lrrr}
\toprule
Model & Rows & Queries & Correct (\%) \\
\midrule
GLM4 & 14,592 & 304 & 37.38 \\
Qwen3 & 14,592 & 304 & 41.18 \\
Gemma4-off & 14,592 & 304 & 43.54 \\
Gemma4-on & 14,592 & 304 & \textbf{46.57} \\
Gemini Flash-0 & 14,592 & 304 & 46.15 \\
Gemini Flash-24k & 14,592 & 304 & 45.96 \\
\bottomrule
\end{tabular}
\caption{Completed common-frame inventory.}
\label{tab:supp-inventory}
\end{table}

With the shared panel and runtime inventory fixed, the next section expands the
main paper's Original-only display to every persona and target. As throughout
the paper, failure to reject a paired difference is not evidence of equivalence
because no equivalence margin was predeclared.

\section{Detailed Persona-Wise Accuracy}

Table~\ref{tab:all-persona-mode-accuracy} complements the Original-phrasing
matrix in the main paper with every persona, target, and visible-CoT
condition. It preserves direct generation as an explicit comparator and shows
the phrasing-dependent cells that are necessarily compressed in the main
narrative.

\begin{table*}[t]
\centering
\begin{tabular}{lllrrrrrrrr}
\toprule
Model & Persona & Target & Direct & \shortstack{NL\\CoT} & \shortstack{SQL\\CoT} & \shortstack{Python\\CoT} & \shortstack{Pseudo\\code CoT} & \shortstack{Length\\filler} & Scrambled & \shortstack{Wrong\\content} \\
\midrule
GLM4 & expert & SQL & 88.16 & 87.83 & 86.18 & 87.83 & 88.82 & \textbf{90.79} & 86.84 & 87.83 \\
GLM4 & expert & PYTHON & 60.20 & 54.61 & 53.95 & 55.59 & 57.89 & 60.20 & \textbf{60.86} & 55.92 \\
GLM4 & novice & SQL & 19.08 & 21.38 & 18.42 & 21.05 & 18.42 & 20.72 & \textbf{26.32} & 20.72 \\
GLM4 & novice & PYTHON & 14.14 & 14.80 & 13.49 & 15.13 & 15.13 & 14.14 & \textbf{16.45} & 13.16 \\
GLM4 & original & SQL & 23.68 & 28.29 & 28.29 & 25.99 & \textbf{29.28} & 24.01 & 25.66 & 27.30 \\
GLM4 & original & PYTHON & 15.13 & \textbf{19.74} & 18.42 & 17.11 & 17.76 & 16.45 & 15.13 & 15.79 \\
\shortstack{Gemini\\Flash-0} & expert & SQL & 92.43 & 91.45 & \textbf{93.42} & \textbf{93.42} & 90.79 & 92.11 & 92.11 & 92.11 \\
\shortstack{Gemini\\Flash-0} & expert & PYTHON & \textbf{68.42} & 65.13 & 67.11 & 64.14 & 67.76 & \textbf{68.42} & 66.45 & 66.78 \\
\shortstack{Gemini\\Flash-0} & novice & SQL & 30.92 & 32.57 & 33.55 & 30.92 & 33.22 & 30.92 & \textbf{38.82} & 34.21 \\
\shortstack{Gemini\\Flash-0} & novice & PYTHON & 20.39 & 22.70 & 23.36 & 20.72 & 22.70 & 21.38 & \textbf{24.01} & 19.41 \\
\shortstack{Gemini\\Flash-0} & original & SQL & 35.86 & 36.51 & 37.17 & 35.53 & 39.80 & 39.47 & 39.14 & \textbf{41.12} \\
\shortstack{Gemini\\Flash-0} & original & PYTHON & 23.03 & 25.00 & 25.99 & 22.04 & 26.32 & \textbf{26.97} & 25.00 & 24.34 \\
\shortstack{Gemini\\Flash-24k} & expert & SQL & \textbf{92.11} & 91.78 & 91.78 & 91.45 & 90.79 & 91.78 & 91.12 & 91.12 \\
\shortstack{Gemini\\Flash-24k} & expert & PYTHON & 66.78 & 66.78 & \textbf{69.08} & 68.75 & 68.42 & 67.43 & 68.75 & 65.79 \\
\shortstack{Gemini\\Flash-24k} & novice & SQL & 31.58 & 31.25 & 32.89 & 30.59 & 30.92 & 31.91 & \textbf{34.54} & 32.57 \\
\shortstack{Gemini\\Flash-24k} & novice & PYTHON & 21.38 & 22.37 & 24.34 & 21.71 & 21.38 & 21.05 & \textbf{25.33} & 19.74 \\
\shortstack{Gemini\\Flash-24k} & original & SQL & 36.84 & \textbf{37.50} & 35.86 & \textbf{37.50} & 36.51 & 36.18 & 36.51 & 33.22 \\
\shortstack{Gemini\\Flash-24k} & original & PYTHON & 26.32 & 25.00 & \textbf{27.63} & 25.99 & 26.32 & 26.64 & \textbf{27.63} & 23.36 \\
Gemma4-off & expert & SQL & \textbf{93.75} & 93.09 & \textbf{93.75} & 92.43 & 93.42 & 92.43 & \textbf{93.75} & \textbf{93.75} \\
Gemma4-off & expert & PYTHON & 64.80 & 64.80 & 64.47 & 64.80 & \textbf{66.12} & 64.14 & 64.80 & 64.47 \\
Gemma4-off & novice & SQL & 29.61 & 30.26 & 28.95 & 30.59 & 29.93 & 27.63 & \textbf{33.22} & 30.92 \\
Gemma4-off & novice & PYTHON & 18.09 & 17.11 & 17.11 & 17.43 & 16.78 & 15.79 & \textbf{18.42} & 17.43 \\
Gemma4-off & original & SQL & 36.18 & 37.17 & \textbf{37.83} & 37.50 & 36.51 & 37.17 & 36.84 & 36.51 \\
Gemma4-off & original & PYTHON & \textbf{19.74} & 17.76 & 19.41 & \textbf{19.74} & 18.09 & 18.42 & 18.42 & 18.75 \\
Gemma4-on & expert & SQL & 92.76 & 92.76 & \textbf{94.41} & 92.76 & 94.08 & \textbf{94.41} & 93.42 & 93.42 \\
Gemma4-on & expert & PYTHON & 65.79 & 66.45 & \textbf{70.07} & 67.11 & 68.09 & 66.12 & 66.45 & 67.76 \\
Gemma4-on & novice & SQL & 30.59 & 31.58 & 30.92 & 30.92 & 31.25 & 29.93 & \textbf{33.22} & 31.25 \\
Gemma4-on & novice & PYTHON & 23.03 & 22.70 & 21.71 & 22.37 & 22.04 & 21.38 & \textbf{24.01} & 21.71 \\
Gemma4-on & original & SQL & 37.83 & 37.83 & 38.16 & 38.49 & 39.14 & 39.14 & 39.47 & \textbf{41.12} \\
Gemma4-on & original & PYTHON & 26.32 & \textbf{27.30} & \textbf{27.63} & 26.32 & 25.66 & 24.34 & 25.99 & 25.99 \\
Qwen3 & expert & SQL & 91.12 & 89.80 & 90.79 & 89.47 & 90.79 & 91.78 & \textbf{92.43} & 91.12 \\
Qwen3 & expert & PYTHON & 58.88 & 58.88 & 59.87 & 54.61 & 58.22 & 61.84 & \textbf{62.83} & 62.17 \\
Qwen3 & novice & SQL & 23.03 & 25.99 & 23.03 & 23.68 & 25.33 & 24.34 & \textbf{29.61} & 25.33 \\
Qwen3 & novice & PYTHON & 16.12 & 19.08 & 16.12 & 13.82 & 17.43 & 18.09 & 18.42 & \textbf{19.74} \\
Qwen3 & original & SQL & 29.93 & 33.55 & 30.26 & 33.55 & 30.59 & 31.58 & \textbf{34.54} & 30.26 \\
Qwen3 & original & PYTHON & 21.71 & 22.04 & 21.38 & 20.39 & 20.72 & 23.03 & \textbf{25.33} & 24.01 \\
\bottomrule
\end{tabular}
\caption{Full persona-wise exact execution accuracy (\%). The best condition
in each model-persona-target row is bold.}
\label{tab:all-persona-mode-accuracy}
\end{table*}

The matrix supplies the descriptive cells for the paired contrasts below; it
does not itself establish a condition effect. In particular, direct generation
remains competitive in multiple rows, so a highlighted control or CoT cell is
not evidence of a universal prompting advantage. The next section specifies
the tests, confidence intervals, multiplicity families, and adjusted covariates.

\section{Paired Statistical Procedures}
\label{sec:appendix-statistics}

\subsection{Notation and analysis unit}

Let $q\in\{1,\ldots,Q\}$ index retained queries, with $Q=304$ in the common
frame. Let $s$ index the six model configurations, $p$ the persona phrasing,
$t$ the target language, and $r$ the reasoning condition. The binary outcome is
\[
  y_{qsptr} =
  \begin{cases}
    1, & \text{correct,}\\
    0, & \text{otherwise.}
  \end{cases}
\]
Here correct means that the generated program executes and matches the gold
result after deterministic normalization.
All paired contrasts hold $q$, $s$, $p$, and $t$ fixed and vary only the
reasoning condition. For the primary family, the comparator $a$ is direct
generation. For the control-mechanism family, $a$ can be direct generation,
NL CoT, another control, or the explicit-CoT mean, depending on the stated
hypothesis.

\subsection{McNemar Test}

For a reasoning condition $r$ and comparator $a$, let
\[
  \begin{split}
  b &= \sum_q I(y_{qsptr}=1,\,y_{qspta}=0),\\
  c &= \sum_q I(y_{qsptr}=0,\,y_{qspta}=1),
  \end{split}
\]
where $I(\cdot)$ is the indicator function. Thus $b$ counts matched queries on
which condition $r$ is correct and comparator $a$ is wrong, while $c$ counts the
opposite discordance. Under the paired null, the $b+c$ discordant outcomes are
exchangeable with probability one half \citep{mcnemar1947note}. Let
$B\sim\mathrm{Binomial}(b+c,0.5)$, $P_-=\Pr[B\le \min(b,c)]$, and
$P_+=\Pr[B\ge \max(b,c)]$. The exact two-sided McNemar/binomial value is
\[
  p_{\mathrm{McN}}=\min\{1,2\min(P_-,P_+)\}.
\]
Concordant pairs do not provide evidence of a difference.

\subsection{Effect Sizes and Confidence Intervals}

The primary effect is
\[
\widehat\Delta_{sptr,a}=\frac{100}{Q}\sum_{q=1}^{Q}
\left(y_{qsptr}-y_{qspta}\right),
\]
reported in percentage points. Here $a$ is the comparator condition and all
other indices are held fixed. We also report a continuity-corrected paired odds
ratio
\[
  \widehat{\mathrm{OR}}_{\mathrm{paired}}=\frac{b+0.5}{c+0.5},
\]
where $b$ and $c$ are the discordant counts defined above. Confidence intervals
resample the $Q=304$ whole-query cluster means with replacement 4,000 times
using seed 20260630 and take the 2.5th and 97.5th bootstrap percentiles.
Resampling individual rows would understate uncertainty because each query
appears in many conditions.

\subsection{BH False-Discovery Control}

If $M$ raw values are ordered $p_{(1)}\le\cdots\le p_{(M)}$, the
Benjamini-Hochberg procedure \citep{benjamini1995controlling} computes the
adjusted value
\[
  p^{\mathrm{BH}}_{(i)}=
  \min_{j\ge i}\left\{\min\left(1,\frac{M}{j}p_{(j)}\right)\right\}.
\]
It controls the expected fraction of false discoveries among rejected
hypotheses, not the probability of any false positive. The 252
mode-versus-direct cells form the primary family. Internal-reasoning
configuration contrasts, target
matching, best-CoT comparisons, persona moderation, and control-mechanism
screens are separate scientific questions and therefore separate declared
families.

\subsection{Token Comparisons}

Prompt, completion, and total-token differences are paired within
query-persona-target, then averaged to one difference per query within each
setting. For a token metric $z\in\{\mathrm{prompt},\mathrm{completion},
\mathrm{total}\}$, the query-level paired token difference is
\[
  d^z_{qsr,a}=\frac{1}{|\mathcal{P}\times\mathcal{T}|}
  \sum_{p,t}\left(z_{qsptr}-z_{qspta}\right),
\]
where $\mathcal{P}$ and $\mathcal{T}$ are the evaluated persona and target sets.
Two-sided Wilcoxon signed-rank tests \citep{wilcoxon1945individual}
operate on those 304 query-level
differences, with BH applied separately to each token metric. Query-cluster
bootstrap intervals target the mean token difference; the rank test targets a
distributional location shift, so the two summaries need not agree for a
heavy-tailed cell. Token use cannot establish whether a prefix was attended to
or semantically ignored.

\subsection{Clustered Logistic GEE}

The complementary \gls{gee} contains 87,552 Bernoulli outcomes and 304 query
clusters. \gls{gee} estimates population-averaged associations under correlated
responses \citep{liang1986longitudinal}. For observation $i$, let $q(i)$ denote
its query cluster and let $\mu_i=\mathrm{E}[y_i]$. We fit a logistic marginal
mean model
\[
  \mathrm{logit}(\mu_i)=
  \beta_0+\beta^\top X_i,
\]
where $X_i$ contains model configuration, reasoning condition, persona, target,
BIRD difficulty \citep{li2023bird}, database, balanced ambiguity score,
phrasing-specificity score,
\gls{scs}, and the declared interaction terms. The estimating equation is
\[
  \sum_{q=1}^{Q} D_q^\top V_q^{-1}(Y_q-\mu_q)=0,
\]
where $Y_q$ is the vector of binary outcomes for query $q$, $\mu_q$ is the
corresponding fitted mean vector, $D_q=\partial\mu_q/\partial\beta$, and $V_q$
is the working covariance matrix. The reported fit uses an independence working
correlation and query-clustered robust sandwich covariance. The exchangeable
working-correlation fit was numerically unstable and failed to converge; it was
discarded. For an omnibus term represented by the contrast matrix $R$, the null
hypothesis is $H_0:R\beta=0$ and the robust Wald statistic is
\[
  W=(R\widehat{\beta})^\top
  \left[R\,\widehat{\mathrm{Var}}(\widehat{\beta})R^\top\right]^{-1}
  (R\widehat{\beta}),
\]
where $\widehat{\beta}$ is the fitted coefficient vector and
$\widehat{\mathrm{Var}}(\widehat{\beta})$ is its query-clustered sandwich
covariance; $W$ is compared with a chi-square distribution having
$\mathrm{rank}(R)$ degrees of freedom. Table~\ref{tab:supp-gee} lists the
BH-adjusted omnibus Wald families.
\begin{table}[t]
\centering
\begin{tabular}{lr}
\toprule
Term family & $p_{\mathrm{BH}}$ \\
\midrule
Persona & $<10^{-6}$ \\
Target & $<10^{-26}$ \\
Model configuration & $<10^{-11}$ \\
Mode & 0.620 \\
Mode $\times$ persona & 0.00033 \\
Mode $\times$ target & 0.626 \\
Model configuration $\times$ mode & 0.0030 \\
Difficulty & 0.179 \\
SCS & 0.362 \\
Ambiguity & 0.0016 \\
Phrasing specificity & 0.0017 \\
\bottomrule
\end{tabular}
\caption{Complementary robust-GEE omnibus tests.  Paired exact contrasts remain
the primary inference.}
\label{tab:supp-gee}
\end{table}

\subsection{Structural Complexity and Other Query Covariates}

\gls{scs} is retained as a standardized query-level structural-complexity
annotation in the adjusted GEE so that a reasoning-condition comparison is not
mistaken for a difference in the structural demands of the query.  Its
BH-adjusted omnibus value in Table~\ref{tab:supp-gee} is 0.362, so it does not
provide a distinct adjusted association with correctness in this panel.
Difficulty is likewise non-significant after adjustment, whereas ambiguity and
phrasing specificity remain associated with correctness.  These variables are
correlated features of the benchmark and persona rewrites, not independently
manipulated causal treatments; the paired condition tests remain the primary
evidence.

These procedures distinguish confirmatory paired comparisons from descriptive
or focused analyses. The next section reports the Original- and Novice-specific
families that were used to probe localized directions without replacing the
primary family.

\section{Original and Novice Focused Families}

Expert phrasing intentionally exposes more task structure, so we also analyzed
Original and Novice phrasings as separate focused families.  These analyses do
not replace the full 252-test primary family.  They ask whether less-explicit
phrasing contains a clearer condition effect after removing expert prompts
from the family under test.
Table~\ref{tab:supp-original-novice-focus} gives the leading pooled focused
contrasts.

\begin{table*}[t]
\centering
\begin{tabular}{lllrrrr}
\toprule
Phrasing & Target & Condition & Direct & Condition & $\Delta$ pp & $p_{\mathrm{BH}}$ \\
\midrule
Original & SQL & Pseudocode CoT & 33.39 & \textbf{35.31} & 1.92 & 0.168 \\
Original & SQL & NL CoT & 33.39 & \textbf{35.14} & 1.75 & 0.168 \\
Original & SQL & Scrambled & 33.39 & \textbf{35.36} & 1.97 & 0.196 \\
Novice & SQL & Scrambled & 27.47 & \textbf{32.62} & 5.15 & 0.00052 \\
Novice & Python & Scrambled & 18.86 & \textbf{21.11} & 2.25 & 0.111 \\
Novice & SQL & NL CoT & 27.47 & \textbf{28.84} & 1.37 & 0.439 \\
\bottomrule
\end{tabular}
\caption{Top pooled condition-versus-direct results in the focused Original
and Novice families.  Pooled contrasts average paired deltas across the six
configurations within each query before testing.}
\label{tab:supp-original-novice-focus}
\end{table*}

Original phrasing contains positive raw movement, especially for SQL-side
visible CoT, but no corrected discovery. Across the 48 Original-only
explicit-CoT per-model cells, the smallest adjusted value is 0.481; across
the 14 pooled Original all-condition tests, the smallest adjusted value is
0.196.  Within the Original-only pooled contrasts, the largest positive SQL-side
movements occur for pseudocode CoT, scrambled context, and NL CoT.

Novice phrasing gives the only focused corrected result, and it is not an
explicit-CoT language: scrambled context improves SQL accuracy by 5.15 points
after correction across the 14 Novice all-condition pooled tests.  Explicit
CoT remains non-significant for Novice phrasing; the smallest adjusted value
in the pooled explicit-CoT family is 0.652.  The narrowed GEE over Original
and Novice rows converges and again does not make the omnibus mode family
significant ($p_{\mathrm{BH}}=0.270$), while the mode-by-phrasing interaction
is suggestive but not corrected ($p_{\mathrm{BH}}=0.109$).

An additional exploratory test averages the four explicit CoT assignments
within query before comparing them with direct generation.  Original-SQL is
+1.56 points (95\% query-bootstrap interval $[0.23,2.93]$, raw $p=0.0239$),
but does not cross the BH threshold across the four focused persona-target
tests ($p_{\mathrm{BH}}=0.0954$).  GLM4 has the clearest model-wise Original
direction: +4.28 points for SQL (raw $p=0.0154$,
$p_{\mathrm{BH}}=0.0921$ across the six models in that persona-target family)
and +3.13 points for Python (raw $p=0.0355$).  No aggregate explicit-CoT cell
is a corrected discovery.  Seventeen of 24 focused model-persona-target
aggregate-CoT deltas are positive; Original-SQL has five positive configurations
and one zero-delta configuration. This is a descriptive fixed-panel direction, not an
independent-systems significance test.  These analyses were motivated after inspection and
are labeled post hoc; they do not change the primary family.

\subsection{Focused Control Mechanism Contrasts}

The focused control analysis keeps Original and Novice phrasing separate and
uses two matched comparators for every ablation.  The NL-CoT comparison asks
whether removing coherent task-relevant content, disrupting order, or
introducing wrong-task content changes the visible-reasoning condition.  The
direct-generation comparison asks whether the same prefix changes accuracy at
all relative to a prompt without visible CoT. Table~\ref{tab:supp-control-mechanisms}
reports the corrected contrasts that localize the Novice-SQL scrambled result.

\begin{table*}[t]
\centering
\begin{tabular}{lllrrrr}
\toprule
Phrasing & Target & Contrast & $\Delta$ pp & 95\% CI & Raw $p$ & $p_{\mathrm{BH}}$ \\
\midrule
Novice & SQL & Scrambled $-$ direct & 5.15 & $[2.80,7.62]$ & $3.72\times10^{-5}$ & 0.00045 \\
Novice & SQL & Scrambled $-$ length filler & 5.04 & $[3.07,7.07]$ & $9.14\times10^{-7}$ & 0.000011 \\
Novice & SQL & Scrambled $-$ wrong content & 3.45 & $[1.54,5.48]$ & 0.00063 & 0.0038 \\
Novice & SQL & Scrambled $-$ explicit-CoT mean & 4.39 & $[2.43,6.41]$ & $1.61\times10^{-5}$ & 0.00019 \\
Novice & Python & Scrambled $-$ length filler & 2.47 & $[0.77,4.22]$ & 0.0046 & 0.0184 \\
Novice & Python & Scrambled $-$ wrong content & 2.58 & $[0.77,4.50]$ & 0.0083 & 0.0250 \\
\bottomrule
\end{tabular}
\caption{Corrected focused control contrasts.  BH is applied within three
separate 12-test families: controls versus direct, between controls, and
controls versus the four-format explicit-CoT mean.}
\label{tab:supp-control-mechanisms}
\end{table*}

The Novice-SQL result is not a generic length effect: no corrected
length-filler difference from direct is detected, while scrambled exceeds both
filler and wrong content.
Nor does it establish a benefit from incoherent reasoning.  The interventions
do not share one frozen trace, and the supplied prefix changes the model's own
generated reasoning behavior.  The most specific supported claim is that the
Novice-SQL configuration is sensitive to a scrambled reasoning-like context.

At the model level, scrambled Novice-SQL gains are +7.89 points for Gemini
Flash-0, +7.24 for GLM4, and +6.58 for Qwen3. These three survive correction
over the 18 controls inside the Novice-SQL screen, but none survives the wider
72-cell focused-control correction.  The pooled fixed-panel result therefore
captures a repeated directional effect within the completed panel.

The difficulty split is shown separately in Table~\ref{tab:supp-scrambled-difficulty}.
Novice scrambled contrasts are positive across all three difficulty levels,
with the strongest corrected evidence on simple and challenging queries and a
smaller direct-generation contrast on moderate queries. The corresponding Original
contrasts do not survive their correction family.  This localization is
consistent with a phrasing-dependent sensitivity, not a difficulty-invariant
property of token permutation.

\begin{table*}[t]
\centering
\begin{tabular}{llrrrr}
\toprule
Persona & Difficulty & $\Delta$ vs NL CoT & $q_{\mathrm{BH}}$ & $\Delta$ vs direct & $q_{\mathrm{BH}}$ \\
\midrule
Original & Simple & 0.00 & 1.0000 & 1.85 & 0.1646 \\
Original & Moderate & -0.62 & 0.8076 & 0.23 & 0.9469 \\
Original & Challenging & 1.31 & 0.4539 & 2.34 & 0.1646 \\
Novice & Simple & 2.85 & 0.0183 & 4.01 & 0.0026 \\
Novice & Moderate & 1.17 & 0.4539 & 3.04 & 0.0183 \\
Novice & Challenging & 3.84 & 0.0077 & 4.12 & 0.0034 \\
\bottomrule
\end{tabular}
\caption{Scrambled-minus-comparator accuracy contrasts by persona and
difficulty, pooled across the six configurations and both targets. Deltas are
percentage points; $q_{\mathrm{BH}}$ values use the focused difficulty family.}
\label{tab:supp-scrambled-difficulty}
\end{table*}

The focused control results motivate a separate look at internal-reasoning
configuration, which is the next channel of variation examined below.

\section{Internal-Reasoning Configuration Detail}

\begin{figure*}[t]
\centering
\includegraphics[width=0.95\textwidth]{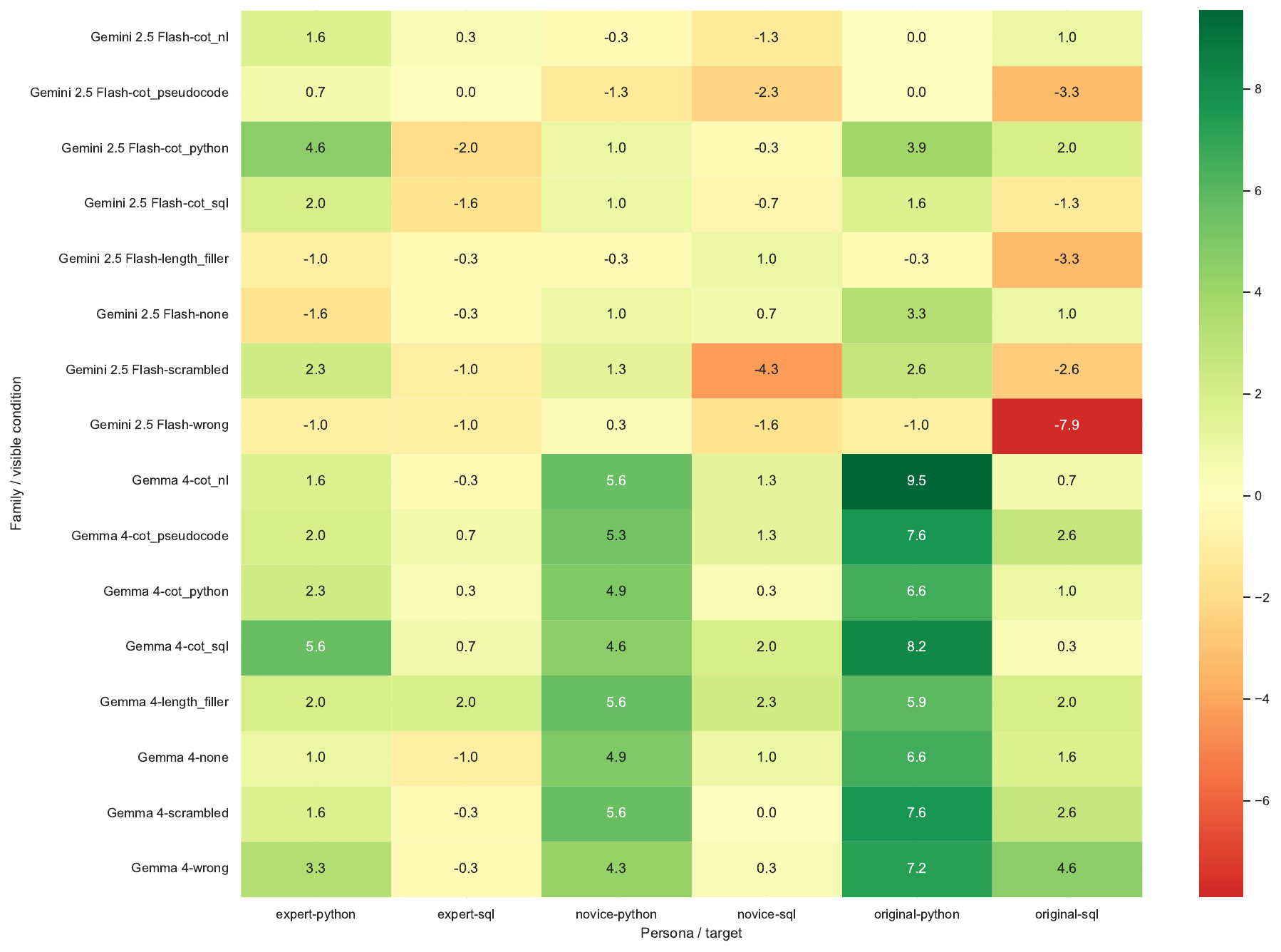}
\caption{Complete same-family internal-reasoning effects (enabled minus
disabled, or budget 24,576 minus 0).}
\label{fig:supp-toggle}
\end{figure*}

The 96-contrast family contains two model families, three personas, two targets,
and eight reasoning conditions. Sixteen cells survive BH. Gemma4 accounts for
15: SQL-CoT under Expert-Python; five Novice-Python visible/control cells; all
eight Original-Python conditions, including direct; and wrong-content
Original-SQL. Gemini Flash contributes one corrected cell: wrong-content Original-SQL,
where Flash-24k is 7.89 points below Flash-0. This distribution explains why a
single pooled ``internal reasoning helps'' coefficient would be misleading.
Figure~\ref{fig:supp-toggle} shows the complete configuration-contrast matrix.

The contrast matrix establishes that changing the internal-reasoning configuration is
not a single global factor. It does not identify the contents or amount of
model-side internal reasoning: the analysis observes only accuracy, resources,
and the externally visible response trace under each configuration. The next
section evaluates whether target-matched explicit CoT or a selected best format
provides a corresponding global rule.

\section{Target Matching, Controls, and Best Formats}

No target-matched comparison survives BH across 36 cells, and no selected best
CoT is significantly superior to every alternative in its corrected family.
This is not evidence that representations are interchangeable; interaction
tests show that their effects vary by persona and model.  It is evidence
against a single global target-matching rule.

Six cells survive the narrower seven-tests-within-cell correction, but none
survives the 252-test primary family. Three are scrambled Novice-SQL gains
(GLM4, Qwen3, and Gemini Flash-0); two are GLM4 Expert-Python losses for NL/SQL
CoT; one is a Gemma4-on Expert-Python SQL-CoT gain. These are localized
within-cell results, not corrected discoveries in the primary family.

Because a representation can change resource use even when it has no global
accuracy advantage, the next section characterizes the associated token and
latency trade-offs.

\section{Efficiency Detail}

The efficiency analysis treats each matched model-persona-target-condition
cell as an accuracy-resource point.  We report direct generation and each of
the four explicit CoTs separately; controls remain in the ablation and
step-compliance analyses but are not assigned a cost/latency comparison here.
This makes the marginal cost of each visible CoT explicit without pretending
that a single accuracy-per-token ratio is comparable across providers.
Table~\ref{tab:supp-efficiency-cot-summary} reports the individual formats,
while Table~\ref{tab:supp-efficiency-family-summary} gives their compact mean
and Table~\ref{tab:supp-efficiency-difficulty-summary} separates task
difficulty from prefix overhead.

\begin{table*}[t]
\centering
\begin{tabular}{lrrrrrr}
\toprule
Reasoning condition & Accuracy (\%) & Mean tokens & Mean latency (s) & $\Delta$ pp & Extra tokens & Extra latency (s) \\
\midrule
Direct & 42.94 & \textbf{2,153} & \textbf{37.1} & 0.00 & 0 & 0.0 \\
NL CoT & 43.41 & 3,095 & 69.3 & 0.48 & 942 & 32.2 \\
SQL CoT & \textbf{43.52} & 3,077 & 75.4 & 0.58 & 924 & 38.4 \\
Python CoT & 42.82 & 3,069 & 76.0 & -0.12 & 916 & 38.9 \\
Pseudocode CoT & 43.51 & 3,194 & 81.3 & 0.58 & 1,041 & 44.2 \\
\bottomrule
\end{tabular}
\caption{Pooled cost and accuracy for each explicit CoT format versus direct
generation over the six configurations, three personas, and two targets. Bold
marks the highest accuracy and the smallest total-token and latency values.}
\label{tab:supp-efficiency-cot-summary}
\end{table*}

\begin{table*}[t]
\centering
\begin{tabular}{lrrrrrrr}
\toprule
Condition family & Accuracy (\%) & Tokens & Latency (s) & $\Delta$ pp & \shortstack{Extra\\tokens} & \shortstack{Extra\\latency (s)} & \shortstack{Dominated\\(T/L, \%)} \\
\midrule
Direct & 42.94 & \textbf{2,153} & \textbf{37.1} & 0.00 & 0 & 0.0 & 0.0 / 0.0 \\
Explicit CoT mean & 43.32 & 3,109 & 75.5 & 0.38 & 956 & 38.4 & 41.7 / 38.9 \\
\bottomrule
\end{tabular}
\caption{Pooled descriptive efficiency summary over six configurations, three
personas, and two targets.  Direct-dominated percentages count matched cells
where direct generation is at least as accurate and no more costly, with one
strict advantage.}
\label{tab:supp-efficiency-family-summary}
\end{table*}

\begin{table*}[t]
\centering
\begin{tabular}{llrrrr}
\toprule
Difficulty & Condition family & Accuracy (\%) & Mean tokens & Mean latency (s) & $\Delta$ pp vs direct \\
\midrule
Simple & Direct & 47.25 & \textbf{1,943} & \textbf{31.9} & 0.00 \\
Simple & Explicit CoT mean & 47.69 & 2,834 & 61.8 & 0.44 \\
Moderate & Direct & 42.52 & \textbf{2,128} & \textbf{34.2} & 0.00 \\
Moderate & Explicit CoT mean & 43.12 & 3,079 & 73.1 & 0.60 \\
Challenging & Direct & 38.20 & \textbf{2,439} & \textbf{46.9} & 0.00 \\
Challenging & Explicit CoT mean & 38.25 & 3,478 & 95.0 & 0.05 \\
\bottomrule
\end{tabular}
\caption{Difficulty-stratified efficiency means pooled across model, persona,
and target cells.  Difficulty raises both baseline and prompted resource use;
the table should not be interpreted as a causal difficulty effect.  From simple
to challenging, direct generation adds approximately 496 tokens and 15.0
seconds, while explicit CoT adds 644 tokens and 33.2 seconds.}
\label{tab:supp-efficiency-difficulty-summary}
\end{table*}

\begin{figure*}[t]
\centering
\includegraphics[width=0.95\textwidth,height=0.80\textheight,keepaspectratio]{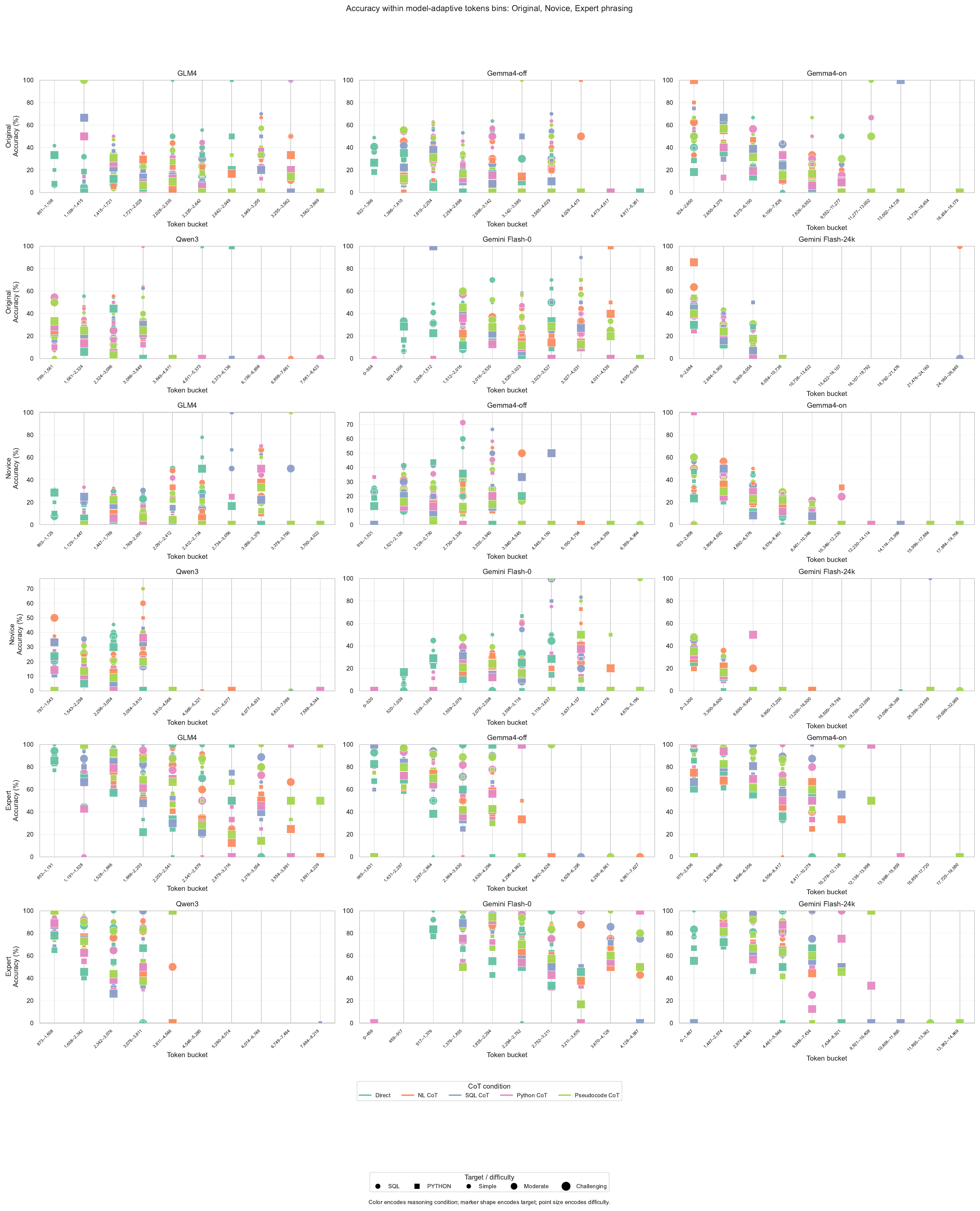}
\caption{Accuracy within model-adaptive token bins, stratified by model,
persona, target, difficulty, and reasoning condition. Original, Novice, and
Expert are shown; point size encodes difficulty and the y-axis is tightened
within each persona row block. Bucket widths target roughly 10\% of the span
and widen when necessary to keep high-range panels readable.}
\label{fig:supp-token-buckets-difficulty}
\end{figure*}

\begin{figure*}[t]
\centering
\includegraphics[width=0.95\textwidth,height=0.80\textheight,keepaspectratio]{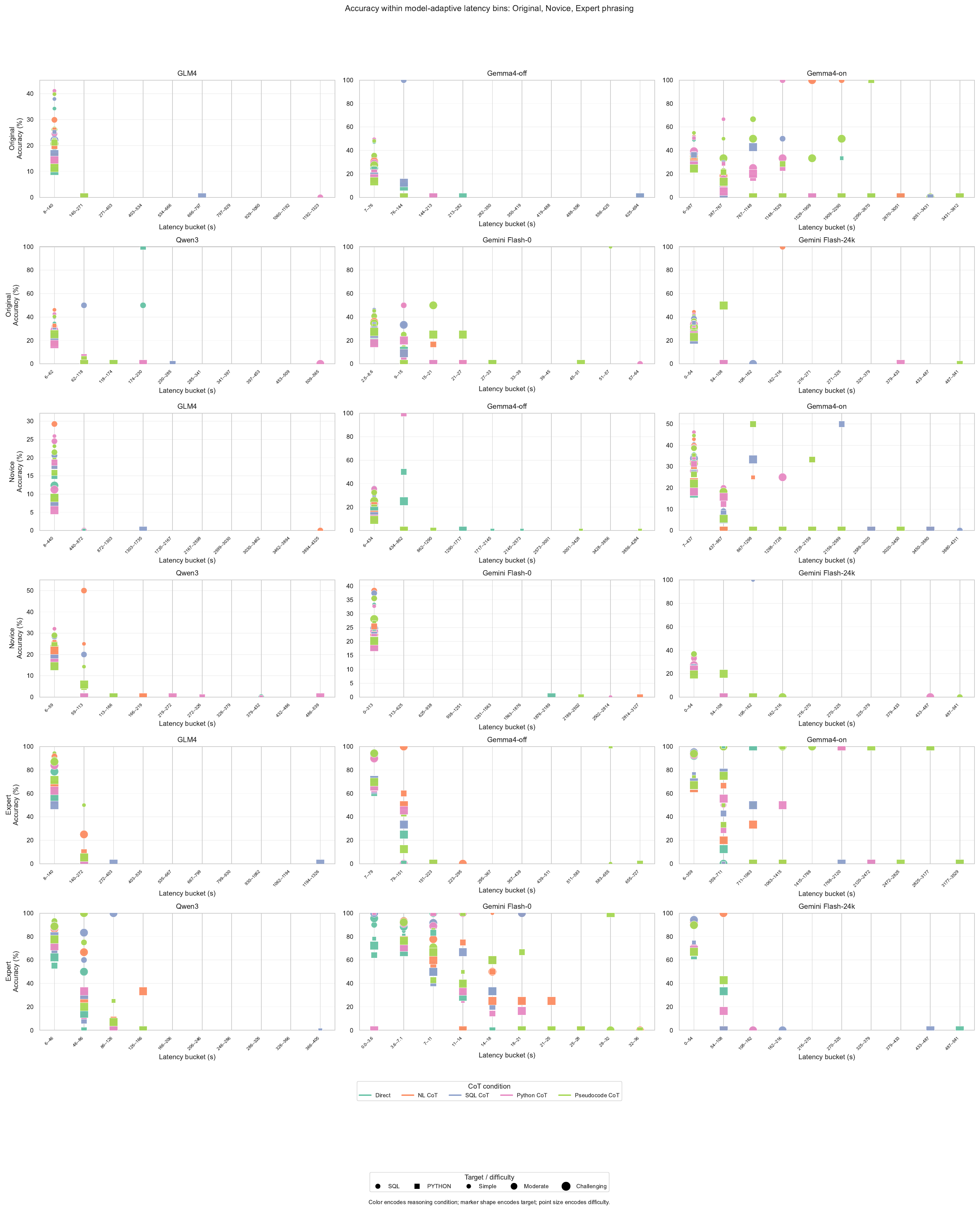}
\caption{Accuracy within model-adaptive latency bins, stratified by model,
persona, target, difficulty, and reasoning condition. Original, Novice, and
Expert are shown; point size encodes difficulty and the y-axis is tightened
within each persona row block. Bucket widths are chosen from roughly 10\% of
the observed span and rounded to the nearest standard latency bin.}
\label{fig:supp-latency-buckets-difficulty}
\end{figure*}

\begin{figure*}[t]
\centering
\includegraphics[width=0.95\textwidth,height=0.80\textheight,keepaspectratio]{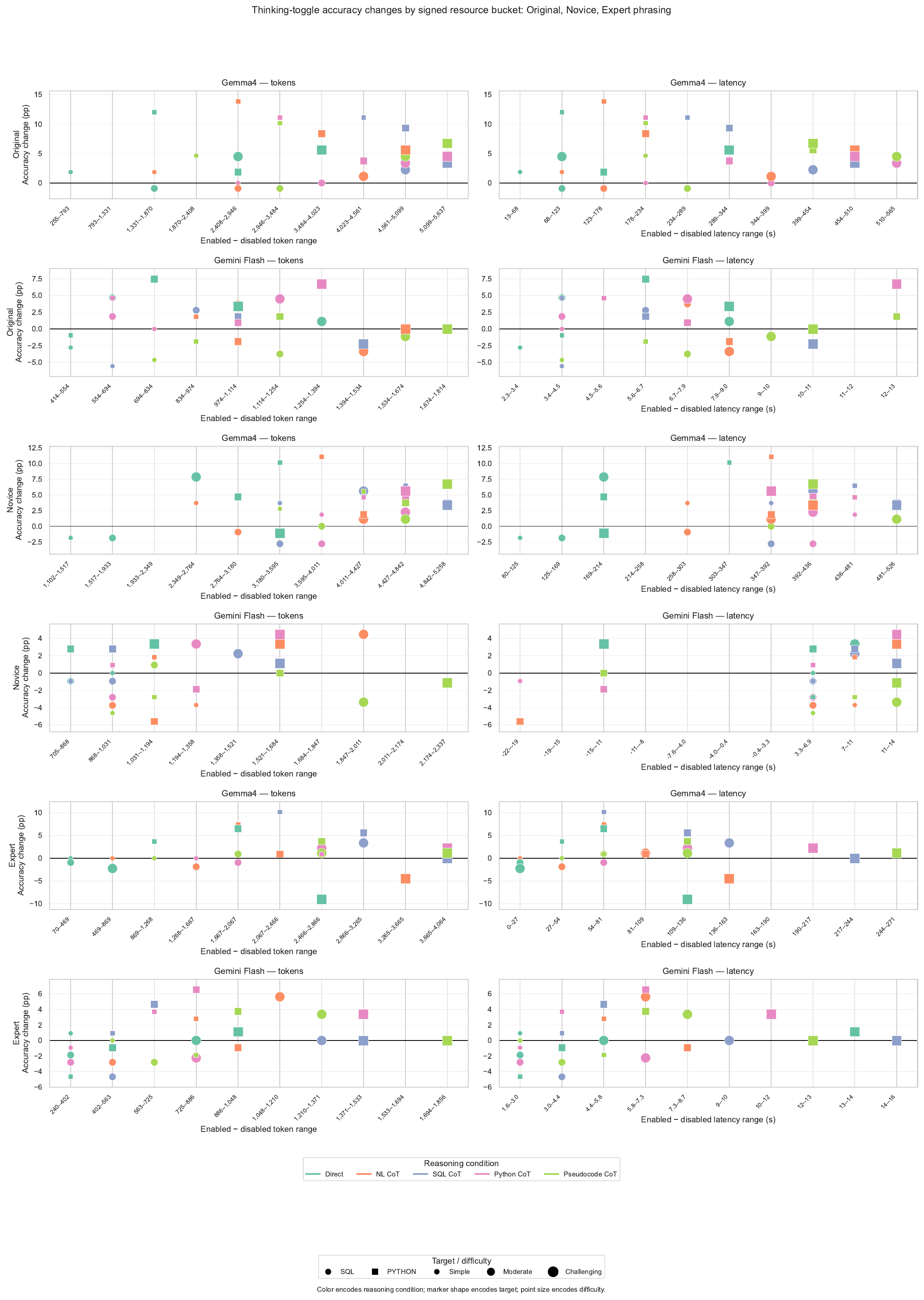}
\caption{Internal-reasoning configuration effects in signed token and latency buckets
for Original, Novice, and Expert phrasing. Color encodes the reasoning condition,
marker shape target, and marker size difficulty.}
\label{fig:supp-thinking-cost-buckets}
\end{figure*}

The bucket tables use model-adaptive bins. The initial width targets roughly
10\% of a model's observed resource span, is rounded to a readable standard
width, and is widened when necessary to retain at most ten plotted bucket
positions in a model panel. Every explicit CoT format remains separate.
Figures~\ref{fig:supp-token-buckets-difficulty} and
\ref{fig:supp-latency-buckets-difficulty} use persona-faceted scatter plots
with the six model configurations arranged in a 2-by-3 layout inside each Original,
Novice, and Expert block. Color encodes the reasoning condition, marker shape
identifies the target language, and point size encodes difficulty; each point
is the mean accuracy within the corresponding model-adaptive bucket.  The
y-axis is tightened separately within each persona row block so that the
central mass of points is easier to read. Bucket widths target roughly 10\% of
the observed span, are rounded to readable standard widths, and are widened for
high-range panels when needed to avoid more than ten plotted positions. Harder tasks
move both direct and CoT conditions toward larger resource budgets, while each
explicit CoT adds further overhead whose accuracy return is model- and
target-dependent.
Figure~\ref{fig:supp-thinking-cost-buckets} applies the same signed-bucket
construction to enabled-minus-disabled thinking changes for each model family.

Table~\ref{tab:supp-efficiency} lists representative accuracy, token, and
latency points used to interpret cost trade-offs.

\begin{table}[t]
\centering
\begin{tabular}{lrrr}
\toprule
\shortstack{Model /\\CoT} & Accuracy & Tokens & Latency (s) \\
\midrule
GLM4 direct & 36.73 & \textbf{1,537} & 20.5 \\
GLM4 NL CoT & 37.77 & 2,222 & 63.8 \\
Qwen3 direct & 40.13 & 1,549 & 13.6 \\
Gemma4-off direct & 43.70 & 1,873 & 28.8 \\
Gemma4-on direct & 46.05 & 3,743 & 144.7 \\
\shortstack{Gemma4-on\\SQL CoT} & \textbf{47.15} & 6,056 & 318.3 \\
\shortstack{Gemini Flash-0\\direct} & 45.18 & 1,742 & \textbf{5.4} \\
\shortstack{Gemini Flash-24k\\direct} & 45.83 & 2,475 & 9.5 \\
\bottomrule
\end{tabular}
\caption{Selected pooled efficiency points. Accuracy is percent; latency is
locally measured client-observed wall-clock time and is not hardware-normalized.}
\label{tab:supp-efficiency}
\end{table}

\begin{figure*}[t]
\centering
\includegraphics[width=0.92\textwidth]{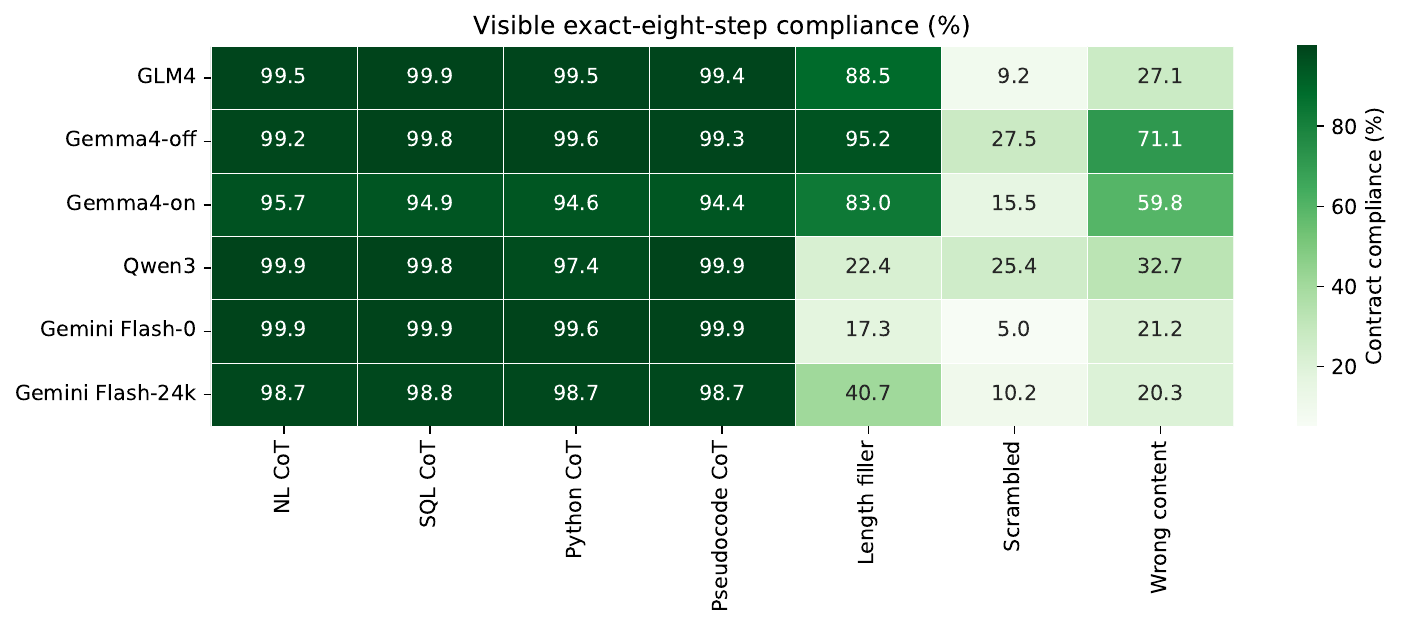}
\caption{Exact eight-step contract compliance for explicit-CoT and control
conditions by model. Low compliance, especially under scrambled context, argues
against interpreting lower completion length as evidence that the control was
simply ignored.}
\label{fig:supp-control-compliance}
\end{figure*}

Every control increases both completion and total tokens relative to direct
generation in all six configurations after BH correction. Compared with the mean of
the four explicit-CoT conditions, control completions are shorter in 14 of 18
model-by-control cells, and all 14 differences are corrected.  Shorter output
does not identify noise rejection: controls still generate 8.6-10.2 thinking
steps on average.  Exact adherence to the intended eight-step count is 58.5\%
for length filler, 15.6\% for scrambled, and 39.2\% for wrong content, compared
with 99.5-99.9\% for explicit CoT.  The control signature is therefore
shorter but more fragmented generated reasoning plus a positive token overhead
relative to direct generation.
Figure~\ref{fig:supp-control-compliance} shows the exact-step compliance
pattern behind this interpretation. Table~\ref{tab:supp-compliance-association}
then tests whether exact-step compliance is associated with correctness within
each control after adjustment for model configuration, persona, target, and
difficulty.

\begin{table}[t]
\centering
\begin{tabular}{lrr}
\toprule
Control & Correctness OR & $q_{\mathrm{BH}}$ \\
\midrule
Length filler & \textbf{1.325} & 0.00048 \\
Scrambled & 0.995 & 0.94542 \\
Wrong content & 1.263 & 0.00264 \\
\bottomrule
\end{tabular}
\caption{Adjusted association between exact eight-step compliance and
correctness within each control. Odds ratios are from query-clustered logistic
GEE; BH correction is across the three control-specific tests.}
\label{tab:supp-compliance-association}
\end{table}

Compliance is positively associated with correctness for length filler and
wrong content, but not for scrambled context
(Table~\ref{tab:supp-compliance-association}). This difference supports two
behavioral axes: adherence to the visible output structure for the labeled
controls, and a separate lexical/context sensitivity after token permutation.
Because compliance is observed rather than randomized, these adjusted
associations do not establish that producing exactly eight steps causes a
correct answer.

The bucket figures should be interpreted as budget-stratified descriptive
comparisons, not causal dose-response curves.  Formal equivalence or
non-inferiority tests with predeclared margins are needed before recommending a
cheaper configuration as a substitute.